\pdfoutput=1

\documentclass[11pt]{article}

\usepackage[preprint]{acl}

\usepackage{times}
\usepackage{latexsym}

\usepackage[T1]{fontenc}

\usepackage[utf8]{inputenc}

\usepackage{microtype}

\usepackage{inconsolata}

\usepackage{graphicx}

\usepackage{microtype}
\usepackage{graphicx}
\usepackage{subfigure}
\usepackage{booktabs}

\usepackage{hyperref}

\usepackage{amssymb}
\usepackage{mathtools}

\usepackage[capitalize,noabbrev]{cleveref}

\usepackage{amsmath}

\makeatletter
\newcommand{\customlabel}[3]{%
   \protected@write \@auxout {}{\string \newlabel {#1}{{#3#2}{\thepage}{#2}{#1}{}} }%
   \hypertarget{#1}{(#2)}
}
\makeatother

\usepackage{caption}

\usepackage{subcaption}

\usepackage{enumitem}

\usepackage{mathabx}

\usepackage{booktabs}
\usepackage{multirow}

\usepackage[acronym,nohypertypes={acronym,notation}]{glossaries}
\newacronym{lm}{LM}{language model}
\newacronym{llm}{LLM}{large language model}
\newacronym{dag}{DAG}{directed acyclic graph}
\newacronym{scm}{SCM}{structural causal model}
\newacronym{sdscm}{SD-SCM}{sequence-driven structural causal model}
\newacronym{nlp}{NLP}{natural language processing}
\newacronym{ite}{ITE}{individual treatment effect}
\newacronym{ate}{ATE}{average treatment effect}
\newacronym{sate}{SATE}{sample average treatment effect}
\newacronym{cate}{CATE}{conditional average treatment effect}
\newacronym{gan}{GAN}{generative adversarial network}

\def \bv{\mathbf{v}}
\def \bV{\mathbf{V}}
\def \bU{\mathbf{U}}
\def \bu{\mathbf{u}}
\def \bx{\mathbf{x}}
\def \bX{\mathbf{X}}

\def \bF{\mathbf{F}}
\def \bY{\mathbf{Y}}
\def \bZ{\mathbf{Z}}
\def \bz{\mathbf{z}}
\def \bPA{\mathbf{PA}}

\def \tv{\tilde{v}}
\def \tu{\tilde{u}}
\def \ty{\tilde{y}}
\def \tx{\tilde{x}}
\def \tt{\tilde{t}}
\def \tg{\tilde{g}}
\def \tm{\tilde{m}}
\def \tw{\tilde{w}}

\def \mG{\mathcal{G}}
\def \mP{\mathcal{P}}

\def \fC{\mathfrak{C}}
\def \fB{\mathfrak{B}}

\def \doI{\text{do}(I)}

\def \doTt{\text{do}(\tilde{t}=t)}
\def \doVi{\text{do}(\tv_i=v)}

\usepackage{xcolor}
\usepackage{tikz}
\usetikzlibrary{shapes,decorations,arrows,calc,arrows.meta,fit,positioning}

\usepackage{float}

\usepackage{amsthm}
\theoremstyle{plain}
\newtheorem{example}{Example}
\newtheorem{theorem}{Theorem}[section]

\theoremstyle{definition}
\newtheorem{definition}[theorem]{Definition}

\theoremstyle{remark}

\usepackage{algorithm}
\usepackage{algpseudocode}
\usepackage[algo2e]{algorithm2e}

\SetKwInput{KwInput}{Inputs}
\SetKwInput{KwReturns}{Returns}
\usepackage{setspace}

\usepackage{wrapfig}

\title{Language Models as Causal Effect Generators}

\author{Lucius E.J. Bynum  \\
New York University \\
  \texttt{lucius@nyu.edu} \\\And
  Kyunghyun Cho \\
  New York University \\
  \texttt{kyunghyun.cho@nyu.edu} \\}

\begin{document}
\maketitle
\begin{abstract}
In this work, we present \glspl{sdscm}, a framework for specifying causal models with user-defined structure and language-model-defined mechanisms. We characterize how an \gls{sdscm} enables sampling from observational, interventional, and counterfactual distributions according to the desired causal structure. 
We then leverage this procedure to propose a new type of benchmark for causal inference methods, generating individual-level counterfactual data to test treatment effect estimation. We create an example benchmark consisting of thousands of datasets, and test a suite of popular estimation methods for average, conditional average, and individual treatment effect estimation. We find under this benchmark that (1) causal methods outperform non-causal methods and that (2) even state-of-the-art methods struggle with individualized effect estimation, suggesting this benchmark captures some inherent difficulties in causal estimation.
Apart from generating data, this same technique can underpin the auditing of language models for (un)desirable causal effects, such as misinformation or discrimination.
We believe \glspl{sdscm} can serve as a useful tool in any application that would benefit from sequential data with controllable causal structure.
\end{abstract}

\section{Introduction}

\begin{figure}[h]
    \centering
    \includegraphics[width=0.48\textwidth]{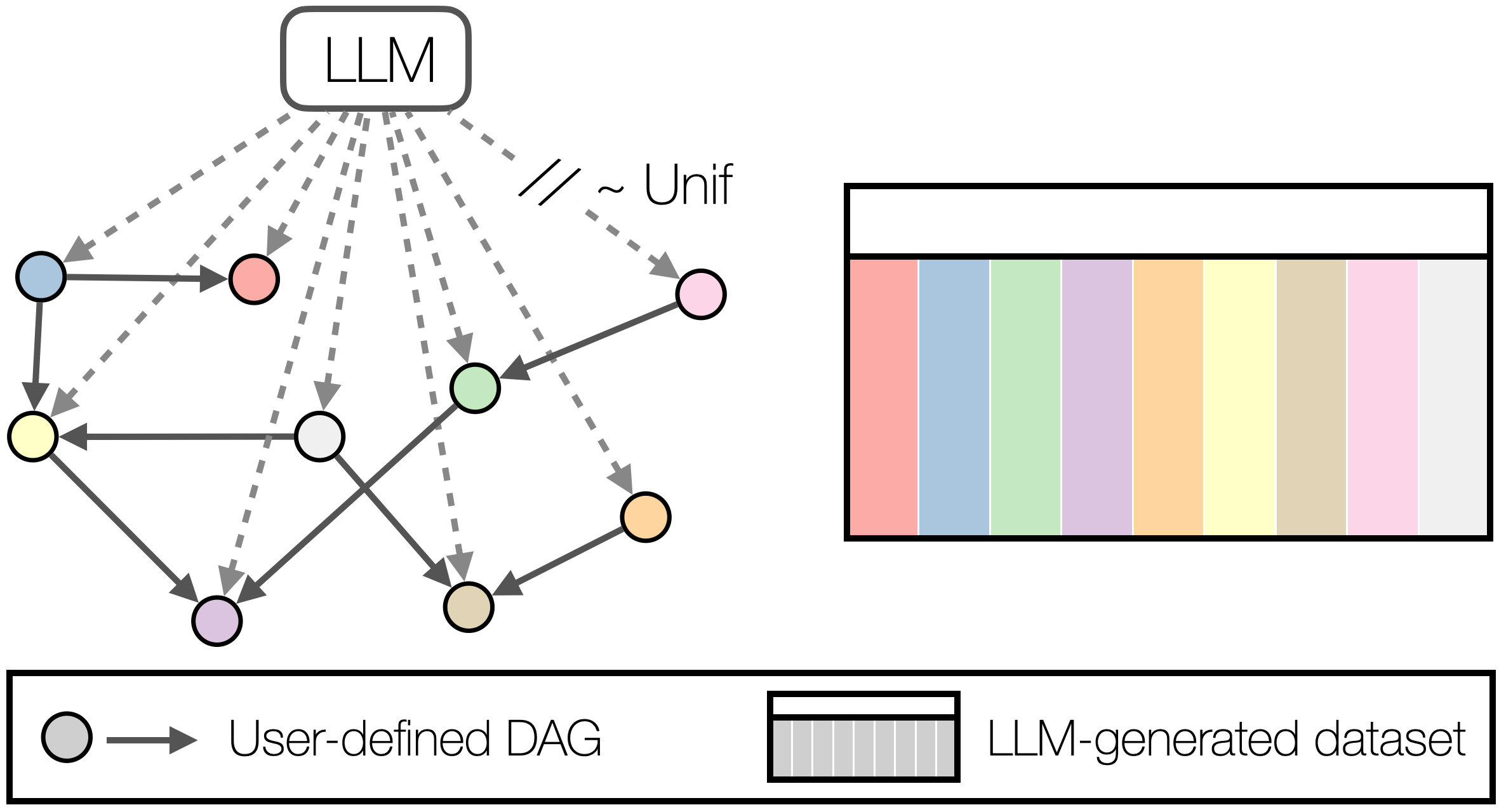}
    \caption{Illustration of a \glsentryfull{sdscm}, which uses a language model to sample data according to a user-specified \glsentryshort{dag}. Any variables whose values are sampled from the language model will potentially share the language model as a common cause (dashed arrows), unless sampled manually, e.g., uniformly.}
    \label{fig:sdscm_diagram}
\end{figure}

Reasoning about counterfactuals plays a fundamental role in understanding cause and effect, both in theory and in practice. Unfortunately, counterfactuals are also fundamentally unobservable \citep{Holland1985StatisticsAC} and must always be simulated. In this work, we leverage \glspl{lm} to help simulate counterfactual data in a user-controlled manner. To achieve this, we borrow the conditional distributions of a pre-trained \gls{lm} in order to parameterize a structural causal model, based on an input \gls{dag} over variables expressed in natural language. This procedure allows us to simulate true counterfactual data --- to observe both potential outcomes --- but, crucially, \emph{without manually specifying functional relationships between variables}. Instead, the specification of structural equations becomes data-driven. We explore how this data-driven approach can enable the specification of causal models for complex settings with less reliance on human expertise or creativity to manually specify relationships between variables. 

\begin{figure*}[h!]
    \centering
    \includegraphics[width=\textwidth]{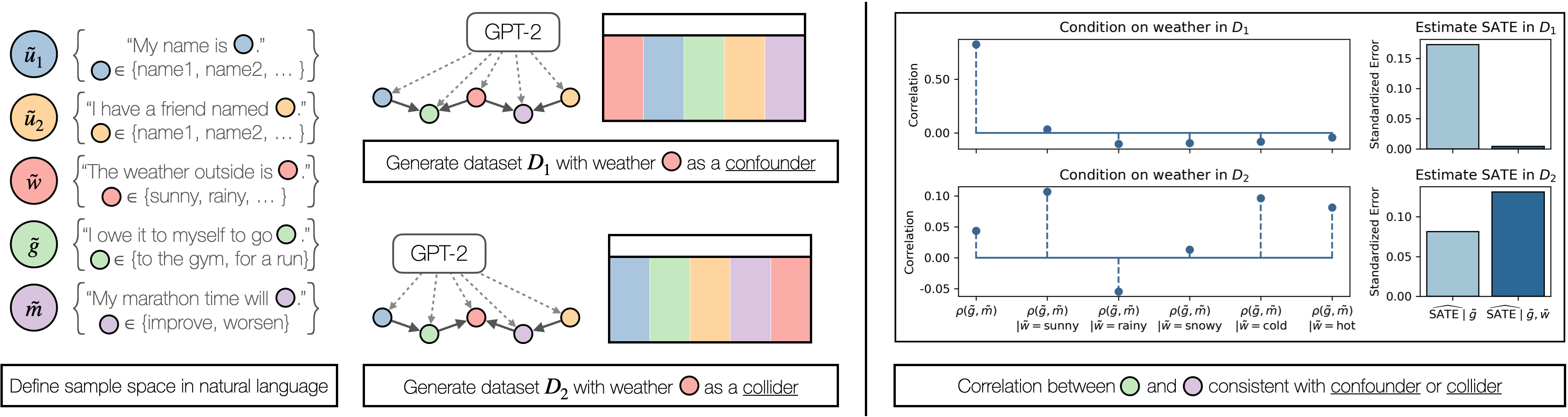}
    \caption{Toy example showing \glspl{sdscm} that use GPT-2 \citep{Radford2019LanguageMA} to generate observational and counterfactual data corresponding to user-specified \glsentryshortpl{dag}. In one case, the red node (weather, $\tw$) is a confounder. In the other case, $\tw$ is a collider. Plots on the right show that despite possible effects from dashed arrows, and \glsentryshortpl{dag} that may contradict what we expect to happen in the real world, the generated data are indeed consistent with $\tw$ as a confounder or a collider.}
    \label{fig:confounder_collider}
\end{figure*}

Many use-cases are possible for sequential data (like text) with controllable causal structure. The main use-case we explore in this work is the development of a new type of benchmark for causal inference --- a benchmark for conditional average and individual treatment effect estimation, where neither the counterfactual outcomes $\ty$ nor the treatment assignments $\tt$ are manually generated. This stands in contrast to existing causal inference benchmarks that must always manually generate $\ty$ or $\tt$, even if covariates are based on real data (see, e.g., \citet{Louizos2017CausalEI}). We find that data generated using our procedure is indeed useful for this task and challenges state-of-the-art estimation methods across both \gls{cate} and \gls{ite}\footnote{It is common to draw no distinction between \glspl{ite} and \glspl{cate} \cite{Vegetabile2021OnTD}, but we emphasize these two quantities as distinct: the \gls{cate} is the conditional expectation of the \gls{ite}, which typically will not explain all \gls{ite} variation \cite{Lei2020ConformalIO}.} estimation.

\subsection{Contribution.} 

\begin{enumerate}[noitemsep]
    \item We define \textbf{a procedure for turning any language model and \gls{dag} into a \acrfull{sdscm}}. \Cref{sec:definitions} characterizes how an \gls{sdscm} provides access to observational, interventional, and counterfactual distributions over sequential data according to the desired \gls{dag}.
    \item In \Cref{sec:benchmark}, we use \glspl{sdscm} to create \textbf{an example benchmark for causal effect estimation} and test a suite of popular estimation methods across \gls{cate} and \gls{ite} estimation. We find that our benchmark challenges state-of-the-art estimation methods. All benchmark datasets as well as code for \gls{sdscm}-based data generation is available on GitHub.\footnote{\url{https://github.com/lbynum/sequence-driven-scms}}
    \item \Cref{sec:auditing} demonstrates how \textbf{this same technique can underpin auditing language models for (un)desirable causal effects}.
\end{enumerate}
Before describing our framework formally, we provide a toy example that illustrates the main points.

\begin{example}[Improving your marathon time at the gym]\label{ex:toy_example}
In this toy example, we use a language model to sample observational and counterfactual data corresponding to two imagined scenarios, each represented by a \gls{dag}. The variables we consider will be represented via sets of sequences, where each set can be viewed as a sample space:
\vspace{-0.3cm}
\begin{itemize}[noitemsep]
    \item $\tu_1$ sample space: ``My name is $x$.'' for all $x \in$ \{John, Jane, Alice, Bob, Charlie\}
    \item $\tu_2$ sample space: ``I have a friend named $x$.'' for all $x \in$ \{John, Jane, Alice, Bob, Charlie\}
    \item \texttt{weather} ($\tw$) sample space: ``The weather outside is $x$.'' for all $x \in$ \{sunny, rainy, snowy, cold, hot\}
    \item \texttt{gymOrRun} ($\tg$) sample space: ``I owe it to myself to go $x$.'' for $x \in$ \{to the gym, for a run outside\}
    \item \texttt{marathonTime} ($\tm$) sample space: ``After this, my marathon time will $x$." for $x \in$ \{improve, worsen\}
\end{itemize}
\vspace{-0.2cm}
The difference between the two scenarios is what we choose to do with the weather variable $\tw$. In the first case, we choose \gls{dag} $\mG_1$ where w is a confounder ($\tu_1 \rightarrow \tg \leftarrow \tw \rightarrow \tm \leftarrow \tu_2$). In the second case, we choose \gls{dag} $\mG_2$ where $\tw$ is instead a collider ($\tu_1 \rightarrow \tg \rightarrow \tw \leftarrow \tm \leftarrow \tu_2$).
Notice we have full control over the \gls{dag} we choose, regardless of what we might expect to happen in the real world or be encoded by the language model (where, for example, we would not expect going to the gym to have any impact on the weather). Each of these \glspl{dag}, by definition, induces a corresponding factorization of the joint distribution across the 5 variables. The factorization for $\mG_1$ is 
$P(\tg | \tw, \tu_1) P(\tm | \tw, \tu_2) P(\tw) P(\tu_1) P(\tu_2)$ 
and the factorization for $\mG_2$ is 
$P(\tw | \tg, \tm) P(\tg | \tu_1) P(\tm | \tu_2) P(\tu_1) P(\tu_2)$.

\textbf{Simulating an observational study:} Our procedure will use a language model to define each of these conditional distributions, instead of defining them manually. In order to observe data that follows the correct structure, we iteratively sample each variable in ancestral order according to the desired \gls{dag}, \textbf{allowing each variable to see only the text of its parents as input}. Doing this allows us to use whatever correlations the language model has encoded to define structural equations. For example, to sample a single observation corresponding to $\mG_1$, the following five phrases (one for each covariate) are sampled using the language model, where [bracketed text] is filled in by querying the model across the corresponding sample space:
\vspace{-0.3cm}
\begin{itemize}[noitemsep]
    \item $\tu_1$ sample: ``My name is [Charlie].''
    \item $\tu_2$ sample: ``I have a friend named [Alice].''
    \item $\tw$ sample: ``The weather outside is [cold].''
    \item $\tg|\tu_1, \tw$ sample: ``My name is Charlie. The weather outside is cold. I owe it to myself to go [to the gym].''
    \item $\tm|\tu_2, \tw$ sample: ``I have a friend named Alice. The weather outside is cold. After this, my marathon time will [improve].''
\end{itemize}
\vspace{-0.3cm}
These five text completions correspond to a single observation, where possible values in each sample space are represented by their index. In other words, we have just observed the data point ($\tu_1$, $\tu_2$, $\tw$, $\tg$, $\tm$) = (4, 2, 3, 0, 0) sampled using \gls{dag} $\mG_1$. \Cref{fig:confounder_collider} shows the result of repeating this process for 1000 observations with $\mG_1$ and 1000 observations with $\mG_2$ using GPT-2 \citep{Radford2019LanguageMA} as the language model. For $\mG_1$, we would expect the magnitude of correlation $\rho$ between $P(\tg=1)$ and $P(\tm=1)$ to decrease if we condition on confounder $\tw$. By contrast, for $\mG_2$, where $\tw$ is instead a collider, we would expect the magnitude of $\rho$ to instead increase if we condition on $\tw$. \Cref{fig:confounder_collider} shows that this is indeed the case --- our sampled data reflects the desired causal structure.

\textbf{Simulating counterfactual data:} We can use a similar procedure to simulate interventions instead of observations: we intervene by manually setting an action (in this case, the value of covariate $\tg$), and \textbf{we create a counterfactual outcome by additionally setting exogenous variables $\tu_1, \tu_2$ and any observed non-descendants of $\tg$ --- $\{\tw\}$ for $\mG_1$ and $\emptyset$ for $\mG_2$}. In \Cref{sec:definitions}, we formally define the correspondence of this process to counterfactual versus interventional distributions. This allows us to directly simulate a counterfactual outcome for each of the observed units, choosing $\tg=1$ or $\tg=0$ during sampling to generate each unit's potential outcomes. We can then, for example, test how well a treatment effect estimation method will perform \textbf{if the estimation method is given only the observational data}, i.e., data without any intervention. The right side of \Cref{fig:confounder_collider} shows prediction error in standard deviation units when using a random forest to predict the \gls{sate} with $P(\tm=1)$ as the outcome, either using treatment $\tg$ as the only covariate, or using both $\tg$ and $\tw$. As we would expect, including a confounder leads to more accurate effect estimation, while including a collider does not. This demonstrates in a simple way the utility of controlled causal data generation --- we can benchmark effect estimation approaches in different settings of interest.

\textbf{Benchmarking \gls{cate} and \gls{ite} estimation:} Many realistic datasets exist for benchmarking estimation of \glspl{ate}, because \glspl{ate} are often feasible to isolate with proper study design. However, there is a lack of such data for benchmarking \gls{cate} and \gls{ite} estimation, where either the outcomes or treatment assignments must always be manually generated. \textbf{The key benefit of simulating data this way is that individual-level counterfactual data are observable and controllable}. This allows us to not only test \gls{ate} estimation methods like in Figure 2, but more importantly to benchmark individual-level effect estimation.
\end{example}
In the remaining sections, we formalize our procedure beyond this toy example and demonstrate how it can be used to generate more complex data that challenges state-of-the-art causal inference methods across both \gls{cate} and \gls{ite} estimation.

\section{Related work} 
\paragraph{Causal inference benchmarks and evaluation.} \citet{Curth2021ReallyDG} lay out four categories of commonly-used methods for semi-synthetic data generation with known causal effects: (1) simulating treatment effects using real baseline outcomes \citep{knaus2021machine}; (2) using real covariates but simulating response surfaces \citep{wendling2018comparing,franklin2014plasmode,Hill2011BayesianNM}; (3) performing biased sampling of randomized data \citep{gentzel2021and,Dehejia1999Causal}; and (4) constructing (proxies of) counterfactuals and interventions from real or empirical data \citep{Louizos2017CausalEI,gentzel2019empirical}. The paradigm of fitting models to real data and then sampling synthetic data from the fit models is common in many works \citep{schuler2017synth,Neal2020RealCauseRC}. In this area, the most closely related works to ours in spirit are those that fit generative models to real datasets such that treatments, outcomes, and covariates --- in effect, entirely new datasets --- can be sampled, such as \citet{athey2024wasserstein} and \citet{Neal2020RealCauseRC}. While such methods are similar in that they rely on generative models, they are fundamentally different from ours, as they are based on individual datasets that already exist (and already have a fixed causal structure), rather than allowing for arbitrary causal structures to be imagined by a user and then parameterized by a generative model. 
Our setup is akin to a high-fidelity simulation environment \citep{mcduff2022causalcity} that provides empirical counterfactual data \citep{gentzel2019empirical}, but without needing to manually design all aspects of the simulation, and in a manner that is instead based on natural language.
This work is also loosely related to methods that parameterize \glspl{scm} with generative models or other deep learning components, such as \citet{pawlowski2020deep,Sanchez2022DiffusionCM}, but such methods are geared towards counterfactual inference and learning causal relationships from existing data, rather than flexible data generation.

\paragraph{Language models and causal inference.} Our work is not the first to suggest that language models can generate outputs that have casual structure.
Many works aim to augment language models with the ability to generate counterfactual data \citep{Chatzi2024CounterfactualTG,Li2023PromptingLL,Betti2023RelevancebasedIF,Hao2021SketchAC,Gat2023FaithfulEO}. Counterfactuals and causal reasoning are useful across various \gls{nlp} tasks, making this capability of particular interest for ongoing \gls{lm} research \citep{Wang2024ASO}, and language models with causal reasoning capabilities have a wide variety of applications both within and beyond \gls{nlp} \citep{Vashishtha2023CausalIU,Jin2023CLadderAB,Zecevic2023CausalPL,Liu2024LargeLM,Feder2021CausalII,Kcman2023CausalRA,Jin2023CanLL,Gat2023FaithfulEO}. 
We are also not the first to point out that counterfactual data generation with language models is useful for understanding the internal `world model' constructed by an \gls{lm} and auditing for bias \citep{Fryer2022FlexibleTG}. The most similar works to ours that we know of are the contemporaneous works \citet{Chatzi2024CounterfactualTG} and \citet{ravfogel2024gumbelcounterfactualgenerationlanguage}, which also model counterfactuals in \glspl{lm} using \glspl{scm}. These works focus on how to generate counterfactual strings after \emph{network interventions} within the \gls{lm} itself. To achieve this, they leverage the Gumbel-Max trick to infer the noise responsible for generating an input and reuse the same noise (or an inferred noise distribution) to generate a corresponding counterfactual output. 
Our work is fundamentally different in two key ways. First, we consider \emph{semantic interventions} rather than network interventions, i.e., modeling causal relationships and counterfactuals all within a semantically meaningful simulation based on a fixed \gls{lm}. Second, we control the causal structure of the data generation process, taking a \gls{dag} as input and generating data according to that \gls{dag}.

In more general terms, \emph{we focus on how to generate data given a desired causal structure}. This capability has important use-cases for downstream tasks like the ones we demonstrate here --- generating treatment effects to benchmark effect estimation methods and testing for encoded effects. But more broadly, we provide a generalization of how sequence data and structural causal models can be combined in order to flexibly generate observational, interventional, and counterfactual data for whatever purpose it might be useful.

\section{Controlled causal data generation via language model}\label{sec:definitions}

In this section, we briefly describe how \glspl{sdscm} enable sampling from observational, interventional, and counterfactual distributions according to the desired causal structure. The full set of definitions, notation, and algorithms for \glspl{sdscm} using structural causal models can be found in \Cref{sec:full_definitions}.

We define a \textbf{sequence variable} $\tx$ as a random variable whose sample space $\Omega_{\tx}$ is a set of sequences. We then define an \gls{sdscm} as a 5-tuple $\fB = (\bV, \bU, \mG, \mP, \tau)$, where $\bV$ is a set of finite-domain endogenous/observed sequence variables and $\bU$ a set of finite-domain exogenous/unobserved sequence variables; $\mG$ is a \gls{dag} over the variables $\tx_i$ in $\bV \cup \bU$ where $\bPA_i \subseteq \left(\bV \cup \bU \right) \setminus \{\tx_i\}$; $\mP$ is a language model trained on prior inputs $C$ whose vocabulary $\mathbb{V}$ contains all tokens used in $\Omega_{\bV}, \Omega_{\bU}$; and $\tau$ is an arbitrary fixed topological ordering of $\bV \cup \bU$ consistent with $\mG$.

The general procedure for sampling data from an \gls{sdscm} relies on two simple ideas: (1) creating concatenated prior inputs for each variable using only the sequences of its parents, which we term \textbf{parent-only concatenation}, and (2) restricting the domain of the \gls{lm} over the current variable's sample space, termed \textbf{domain-restricted sampling}. Sampling proceeds in topological order according to $\tau$, which is required in order to break ties between parents, since \glspl{lm} can be sensitive to even small changes in phrasing. The key difference between an \gls{scm} and an \gls{sdscm} is that all variables have at least one common ancestor --- the prior inputs $C$ that were used to train the language model.\footnote{It would also be possible to \emph{train} an \gls{lm} to induce distributions over the desired variables given this setup, which we leave to future work.}

\paragraph{Observational samples.} Observational data are sampled using parent-only concatenation and domain-restricted sampling for each variable according to $\tau$ (\Cref{sec:sdscm}). 

\paragraph{Interventional samples.} The sequence-driven interventional distribution, given $\doVi$ as the intervention, samples data in the same manner as observational sampling, but now with variable $\tv_i$ replaced by value $v$ during sampling (\Cref{def:sdscm_interventional_distribution}).

\noindent \begin{minipage}{0.48\textwidth}
\begin{algorithm}[H]
\setstretch{1.1}
\caption{A single \gls{sdscm} sample from the counterfactual distribution given observation $\mathbf{s}_{\textrm{obs}}$}\label[algorithm]{alg:counterfactual_sampling}
\KwInput{$\mathbf{s}_{\textrm{obs}} = \left(u_1, \ldots, u_{|\bU|}, v_1, \ldots, v_{|\bV|}\right) \newline \doVi, \fB = (\bV, \bU, \mG, \mP(\cdot), \tau)$}
\KwReturns{$\mathbf{s}^* = \left(u_1, \ldots, u_{|\bU|}, v^*_1, \ldots, v^*_{|\bV|}\right)$}

$\mathbf{s}^* \gets (u_1, \ldots, u_{|\bU|})$

$\textrm{ND}_i \gets$ non-descendants of $\tv_i$ in $\mG$

\For{$\tx_t \in \tau \setminus \bU$}{
    \If{$\tx_t \equiv \tv_i$}{
        $x_t \gets v$
    }
    \ElseIf{$\tx_t \in \textrm{ND}_i$}{
        $x_t \gets \mathbf{s}_{\textrm{obs}}[t]$
    }
    \Else{
        $\textrm{PA}_\tau \gets \{t' : \tx_{t'} \in \textrm{PA}_{\tx_t}\}$ ordered by $\tau$
        
        $x_{\textrm{PA}_\tau} \gets \bigoplus_{x \in \mathbf{s}^*\left[\textrm{PA}_\tau \right]} x$
        
        $\mathbf{p}_{\tx_t} \gets []$
        
        \For{$k \in 1, \ldots, \left|\Omega_{\tx_t}\right|$}{
            $x \gets \Omega_{\tx_t}[k]$
            
            $\mathbf{p}_{\tx_t}[k] \gets \mP\left( x_{\textrm{PA}_\tau} \oplus x\right)$
        }
        $P_{\textrm{tot}} \gets \sum_k \mathbf{p}_{\tx_t}[k]$
        
        $j \sim \textrm{Multinomial}(\mathbf{p}_{\tx_t} / P_{\textrm{tot}})$
        
        $x_t \gets \Omega_{\tx_t}[j]$
    }
    $\textrm{append}(\mathbf{s}^*, x_t)$
}
\textbf{return} $\mathbf{s}^*$
\end{algorithm}
\end{minipage}

\paragraph{Counterfactual samples.} Counterfactual samples require some additional steps. In order to admit unique answers to counterfactual queries, we define abduction for an \gls{sdscm} given evidence $\bZ=\bz$ as the setting of values $\bU=\bu$ as well as any evidence in $\bZ$ upstream of the intervention. In order to obtain such values $\bu$, one needs access to more than just the endogenous variables $\bV$ and language model $\mP$ --- obtaining $\bu$ requires performing bookkeeping \emph{during the data generation process}.\footnote{This is a restatement of the fact that computing point counterfactuals in \glspl{scm} requires causal mechanisms that are invertible with respect to the noise variables in order to uniquely reconstruct the noise that produced a given observation.} Because our primary application of \glspl{sdscm} in this work is data \emph{generation}, such bookkeeping is possible in all our use cases. \Cref{alg:counterfactual_sampling} shows our procedure for sampling a counterfactual for intervention $\doVi$ given observed unit $\mathbf{s}_{\textrm{obs}} = \left(u_1, \ldots, u_{|\bU|}, v_1, \ldots, v_{|\bV|}\right)$ (see also \Cref{def:sdscm_counterfactual_distribution} for additional discussion).

\begin{figure*}    
    \centering
    \begin{minipage}{0.17\textwidth}
        \centering
        \includegraphics[width=\textwidth]{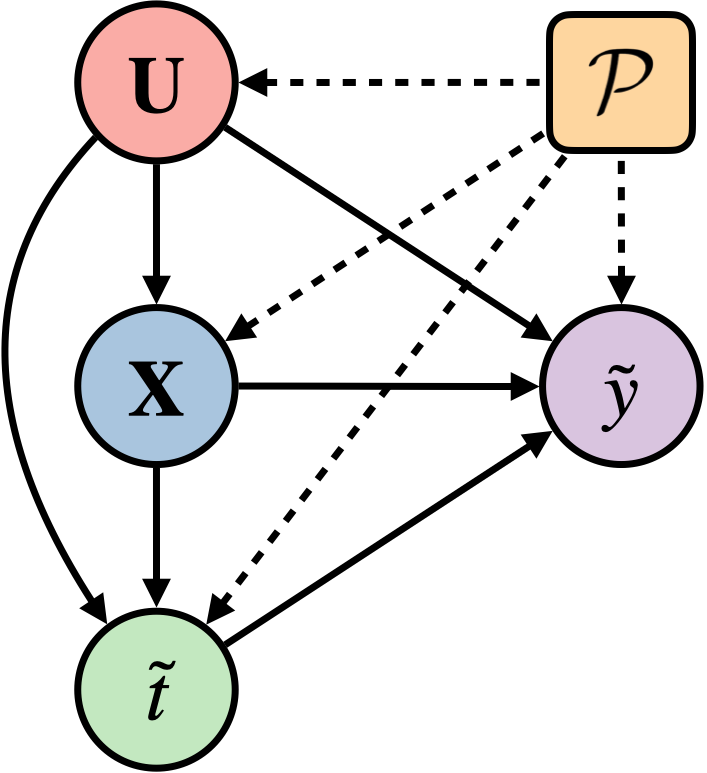}\\
        \customlabel{fig:benchmark_dag}{a}{\ref{fig:benchmark_setup}}
    \end{minipage}
    \begin{minipage}{0.81\textwidth}
        \centering
        \includegraphics[width=\textwidth]{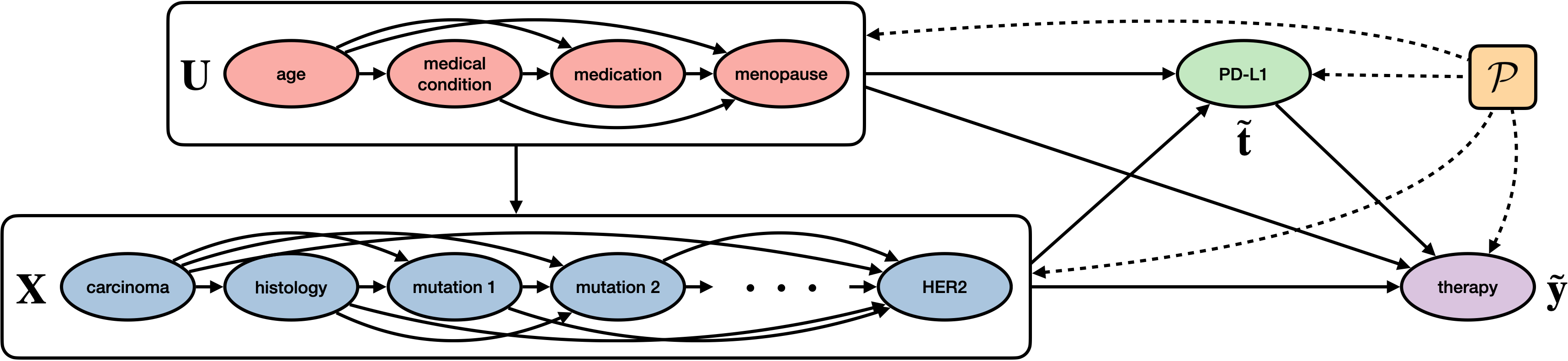}\\
        \customlabel{fig:bcancer_dag}{b}{\ref{fig:benchmark_setup}}
    \end{minipage}
    \caption{\textbf{(a)} A useful \gls{dag} structure for an \gls{sdscm}-generated estimation benchmark. \textbf{(b)} Visual depiction of the structure in (a) used to create the breast cancer \gls{sdscm} in \Cref{sec:bcancer_sdscm}.}
    \label{fig:benchmark_setup}
\end{figure*}

\section{Generating a benchmark for causal effect estimation}\label{sec:benchmark}

To design an \gls{sdscm}-generated benchmark, we focus on the fully sequential \gls{dag} structure shown in Figure~\ref{fig:benchmark_dag}. Exogenous variables $\bU$ precede covariates $\bX$, which in turn precede treatment $\tt$. All variables precede outcome $\ty$. Recall that the presence of an edge in a \gls{dag} allows for \emph{the possibility of a relationship}, but it is the structural equations that determine whether or not a given relationship is meaningful. The strongest assumptions encoded by a \gls{dag}, then, are those edges that are \emph{not present}. Our goal here is to have a language model $\mP$ make as many `decisions' about the data generating process as possible. We thus choose this fully-connected structure as a means of letting $\mP$ define whichever structural equations are meaningful or not given a topological order, and focus on the edge $\tt \rightarrow \ty$ as the target for effect estimation. The key criterion we consider for a useful benchmark is that the datasets we generate require the use of causal reasoning (e.g., controlling for confounding) to recover the effect of $\tt$ on $\ty$. Specifically, we aim to generate data for which the observational and interventional distributions are different, i.e., $P^{\fB}_{\ty \mid \tt=t} \neq P_{\ty}^{\fB; \doTt}$. This criterion is not directly in our control given a fixed language model $\mP$.\footnote{We discuss applications of this same idea to \emph{training} a model in \Cref{sec:conclusion}.} However, even with fixed $\mP$ and a fixed \gls{dag}, we find we are able to achieve it through our choice of sample spaces $\Omega_{\bU}$, $\Omega_{\bX}$, $\Omega_{\tt}$, $\Omega_{\ty}$.

\subsection{Breast cancer \glsentryshort{sdscm}}\label{sec:bcancer_sdscm}

We define an \gls{sdscm} over 14 variables in order to explore the effect of a tumor's PD-L1 expression levels on different breast cancer therapy plans. Our goal with this \gls{sdscm} is to induce causal structure that can challenge estimation algorithms. Covariates in the breast cancer \glspl{sdscm} are defined in full detail in \Cref{appendix:full_bcancer_description} and correspond to the \gls{dag} in Figure~\ref{fig:bcancer_dag}. For each covariate, 10 different phrasings are considered, resulting in a sample space of $10^{14}$ possible sequences.\footnote{Language models are also used to generate the phrasings, but we leave full automation of this process to future work.} We consider 50 different \gls{sdscm} variations, where the sample space for a given \gls{sdscm} is defined by choosing a randomly sampled phrasing from among the possible phrasings for each of the 14 covariates. Then, for each of the 50 \glspl{sdscm}, 20 datasets of size 1,000 are sampled, for a total of 1,000 datasets per language model. We show results for GPT-2 \cite{Radford2019LanguageMA} and Llama-3-8b \cite{Dubey2024Llama}, but we emphasize that the language model is a fully modular component, and thus other language models can be used. For the results shown here, we use $\text{logP}(\ty = 0)$ as the outcome, as there are frequently individual-level effects in probability space. See \Cref{appendix:full_bcancer_description} for example plots of features, propensity scores, and \gls{ite} distributions of the generated data using different possible outcomes. We find that how similar $P_{\ty}^{\fB; \doTt}$ is to $P_{\ty \mid \tt=t}^{\fB}$ varies across \glspl{sdscm}, which we explore further by comparing the performance of observational versus casual estimation approaches.

\subsection{Effect estimation results}

\begin{figure*}    
    \centering
    \begin{minipage}{0.495\textwidth}
        \centering
        \includegraphics[width=\textwidth]{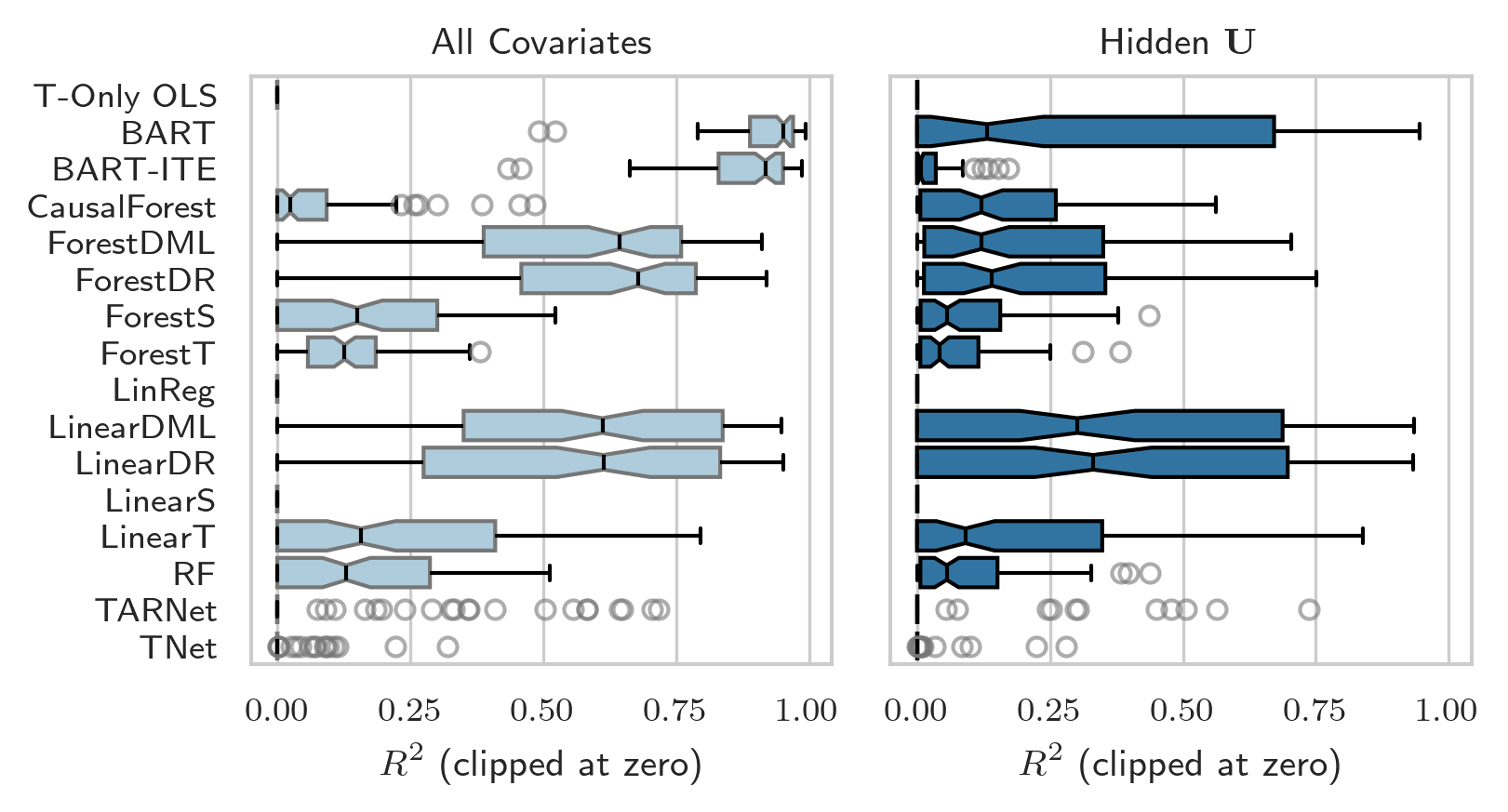}\\
        \customlabel{fig:r2_results}{a}{\ref{fig:cate_results}}
    \end{minipage}
    \begin{minipage}{0.495\textwidth}
        \centering
        \includegraphics[width=\textwidth]{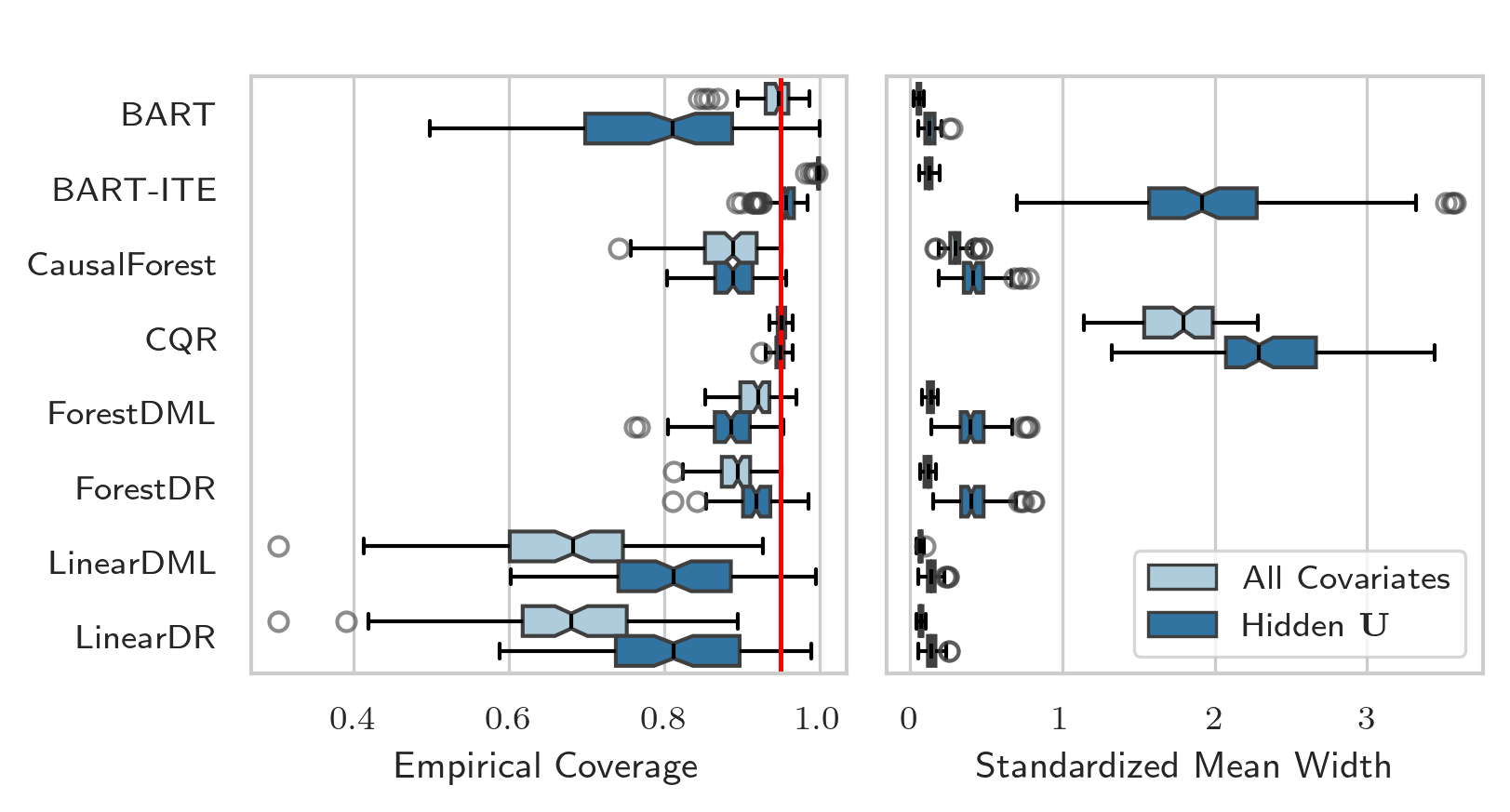}\\
        \customlabel{fig:coverage_results}{b}{\ref{fig:cate_results}}
    \end{minipage}
    \caption{\gls{cate} and \gls{ite} estimation on \gls{sdscm} datasets of size 10,000 generated using Llama-3-8b. \textbf{(a)} $R^2$ values across all methods that provide point estimates. \textbf{(b)} Empirical coverage ($\alpha = 0.05$) and interval width (in outcome standard deviation units) for methods that provide intervals. Nominal coverage of 95\% is indicated by the red line.}
    \label{fig:cate_results}
\end{figure*}

\begin{figure}
    \centering
    \includegraphics[width=0.48\textwidth]{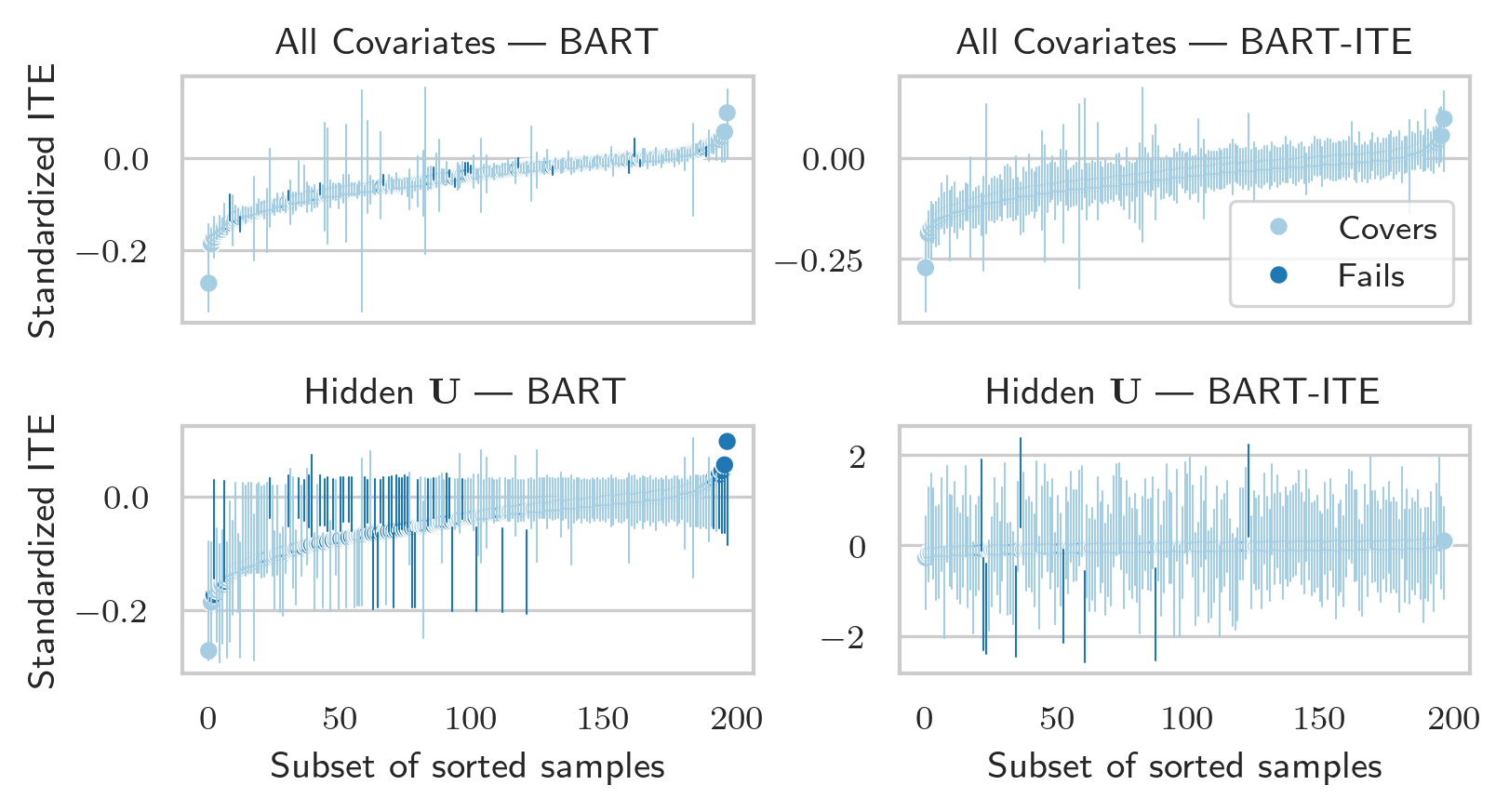}
    \caption{Interval estimates of \glspl{cate}/\glspl{ite} from BART versus BART-ITE on an example \gls{sdscm} dataset.}
    \label{fig:ite_intervals_10k}
\end{figure}

We compare the performance of several effect estimation algorithms. As a naive baseline, ordinary least squares using only the treatment $\tt$ is considered (\textbf{T-Only OLS}). Against this baseline, we consider several causal inference methods of different types, including the causal forest \cite{Wager2018Estimation,Athey2019Generalized} (\textbf{CausalForest}) and two double machine learning methods for \gls{cate} estimation, one linear (\textbf{LinearDML}) and one non-parametric (\textbf{ForestDML}) \cite{chetverikov2016double,Athey2019Generalized,nie2021quasi,chernozhukov2017orthogonal,foster2023orthogonal,mackey2018orthogonal,econml}. We also include two doubly robust meta-learning methods \cite{kunzel2019metalearners}, again, one linear (\textbf{LinearDR}) and one non-parametric (\textbf{ForestDR}), and add Bayesian additive regression trees (\textbf{BART}) \cite{Hill2011BayesianNM,Chipman2008BARTBA} as a widely-used Bayesian non-parametric example.
To represent simpler methods we include linear and non-parametric S- and T-learners (\textbf{LinearS}, \textbf{LinearT}, \textbf{ForestS}, \textbf{ForestT}).
As points of reference for NN-based \gls{cate} estimation methods, we include an NN-based T-learner (\textbf{TNet}), and the NN-based \textbf{TARNet} \cite{shalit2017estimating}. Additional baselines include a random forest baseline (\textbf{RF}) that fits a single response surface and directly predicts treatment effects for each unit, and a linear regression baseline (\textbf{LinReg}) that takes the conditional mean difference (the fit coefficient on $\tt$) to be the effect. Finally, we include two methods that target \glspl{ite} specifically. One method uses BART posterior draws specifically for \glspl{ite} instead of \glspl{cate} (\textbf{BART-ITE}), and the other is conformalized counterfactual quantile regression (\textbf{CQR}) \cite{Lei2020ConformalIO}, which provides conformal inference-based interval estimates of \glspl{ite}. 

\emph{All methods are fit using the default settings of their publicly-available implementations}.\footnote{The causal forest, DML, and DR implementations are provided by \cite{econml}, the BART methods by \cite{Dorie2020CausalIU}, the NN-based methods by \cite{Curth2021ReallyDG,curth2021inductive,curth2021nonparametric} and CQR by \cite{Lei2020ConformalIO}. All code to reproduce this benchmark is available at \url{https://github.com/lbynum/sequence-driven-scms}.} While additional hyperparameter tuning, etc. could be performed for several methods on a case-by-case basis, this section demonstrates what estimation results we get off-the-shelf.

\paragraph{A note on identification.}
Estimation algorithms are designed to work when identification assumptions are met, many of which are untestable. In this section, we demonstrate how \glspl{sdscm} can provide a playground to empirically test how algorithms perform not only in ideal conditions but also when untestable assumptions are not met. This is particularly relevant for \glspl{cate} and \glspl{ite}, where, for example, we might not expect to measure all relevant covariates for each individual unit. In other words, in practice we might not expect to satisfy ignorability. We consider two settings to explore this question empirically. In the `All Covariates' setting, all 14 covariates are observed. In `Hidden $\bU$,' $\bU$ $=\{\tu_1, \tu_2, \tu_3, \tu_4\} =$ \{age, medical conditions, medication, menopausal status\} is hidden.

\subsubsection{Average treatment effects}\label{sec:ate_estimation}

Though we focus on \gls{cate} and \gls{ite} estimation, we first confirm in \Cref{appendix:ate_results} that methods can recover the \gls{ate}. We find that (1) there is a meaningful gap between casual and observational methods and that (2) estimation performance does indeed drop significantly when $\bU$ is hidden.

\subsubsection{\Glsentryshort{cate} and \glsentryshort{ite} estimation}\label{sec:cate_ite_estimation}

To lessen the impact of finite-sample issues, we test on datasets of size 10,000, aggregated within each \gls{sdscm} variation. We show results for Llama-3-8b-generated data in this section, but find similar trends with GPT-2 as well as with dataset size 1,000 in \Cref{appendix:additional_results}. Figure~\ref{fig:r2_results} shows $R^2$ values clipped at zero across all methods that provide point estimates for \glspl{cate}.  When all covariates are observed, BART explains the most \gls{cate} variation, while DML and DR methods do as well at times but with a much lower average. However, \gls{cate} estimation becomes much more challenging for all methods with hidden $\bU$, where no methods perform well. \Cref{fig:pehe_boxplots_10k,fig:pehe_boxplots_sd_ite_10k} in \Cref{appendix:additional_results} show the same results in terms of PEHE (Precision in Estimating Heterogeneous Effects) \cite{Hill2011BayesianNM}, revealing that with no clipping, BART-ITE shows large outliers with hidden $\bU$ and NN methods show large outliers in both settings.

When the \gls{ite} varies due to covariates not conditioned on in the \gls{cate}, as in the hidden $\bU$ setting, the two quantities are distinct. In such cases, uncertainty is especially important. Figure~\ref{fig:coverage_results} shows empirical coverage results for all estimators that provide intervals. With all covariates, empirical coverage is under nominal for all methods that target \gls{cate}, except BART. Hidden $\bU$ increases uncertainty, but also brings coverage closer to nominal for several methods, like LinearDML and both DR methods. CQR remains at or above nominal coverage, but with much wider intervals, as does BART-ITE in the \gls{ite} setting. \Cref{fig:ite_intervals_10k} demonstrates this further, comparing intervals for BART targeting CATE versus BART-ITE on an example dataset. With all covariates (top row), intervals from either method are informative about individual-level effects. However, under hidden $\bU$ (bottom row), the tighter intervals of BART targeting the CATE are overconfident with variable coverage, and the wider intervals of BART-ITE are so wide as to be vacuous, even if we want just a ranking of the \glspl{ite}.

\paragraph{Takeaways.} We summarize a few takeaways for off-the-shelf estimation performance in this example. The first is that \textbf{linear and tree-based methods are often able to perform well}. Second, in a real-world setting, where a method might often be used with its default parameters, \textbf{stability can be important} (e.g., the NN method performance suffers often due to lack of stability). The third takeaway is that \textbf{hidden confounding has a big impact}, across all methods. Even methods that perform particularly well with all covariates (like BART) suffer significantly under hidden $\bU$. Finally, \textbf{\gls{ite} intervals can be unstable and/or vacuous for decision making}, especially with hidden variables, and should thus be used carefully.

\section{Auditing language models for (un)desirable causal effects}\label{sec:auditing}

\begin{figure}
    \centering
    \includegraphics[width=0.48\textwidth]{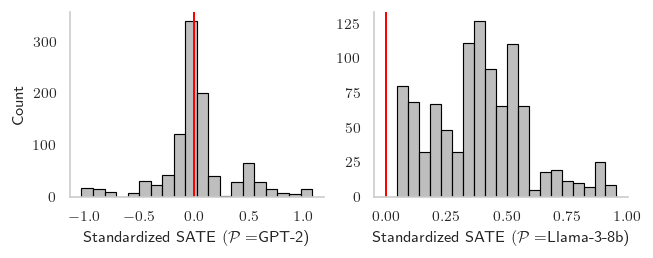}
    \caption{\glspl{sate} across all breast cancer \glspl{sdscm} for outcome $P(\texttt{therapy}=2)$.}
    \label{fig:sate_audit}
\end{figure}

The same framework we use to generate causal effects and benchmark effect estimation methods can allow us to inspect what causal information has been encoded semantically in an \gls{lm}. For example, we can ask, \emph{``Given a world-view described by an input \gls{dag}, what causal conclusion is implied by the language model?''} Our collection of breast cancer \glspl{sdscm} is already set up to explore the effect of PD-L1 on chosen therapy plans, while allowing us to marginalize out an important source of variability: phrasing. Essentially, this amounts to reverse engineering the decision-making process of clinicians, as learned from whatever data the language model was trained on. 

\Cref{fig:sate_audit} shows one example where the two language models \emph{strongly disagree} on what the causal effect is. The effect in this case is the change in probability of choosing the second therapy plan, ``start a regimen of trastuzumab and pertuzumab'' (shown in standard deviation units). GPT-2 has encoded that on average, an increase in PD-L1 expression levels has neither a positive nor negative impact on choosing this therapy plan. However, Llama-3-8b has encoded instead that an increase in PD-L1 always increases the likelihood of this therapy plan. This discrepancy indicates that \textbf{these two language models have encoded two meaningfully different causal effects}. We believe the same procedure can underpin more thorough auditing of \glspl{lm} for misinformation or discrimination, enabling, e.g., path-specific counterfactual fairness analysis \cite{Kusner2017CounterfactualF,Chiappa2018PathSpecificCF}.

\section{Conclusion and Future Work}\label{sec:conclusion}

In this work, we have introduced \acrfullpl{sdscm} as a framework for specifying \glspl{scm} with user-defined structure and \gls{lm}-defined mechanisms. We demonstrate an important use-case for \glspl{sdscm} by creating a benchmark for causal effect estimation. In this proof of concept, we focused on estimation in the presence of confounding, but there are many other settings to explore for effect estimation, such as instrumental variables \cite{Angrist1993IdentificationOC,Hernn2006InstrumentsFC}. 
Using \glspl{sdscm} to additionally test \emph{causal discovery} is of immediate interest, for example, allowing us to test whether a structure learning method can identify whether one variable is causally upstream or downstream of another \cite{Krmer2013CausalAA}. Another significant area of future work is to use \glspl{sdscm} or similar as a means of specifying causal structure over sequential data \emph{during learning} \cite{Im2024UsingDA}; rather than use pre-trained \glspl{lm} to generate effects, a model can be trained or fine-tuned to handle tasks that require causal reasoning, including complex confounding and sequential decision making. 

In short, we believe \glspl{sdscm} can serve as a stepping stone for any application that would benefit from sequential data with controllable causal structure.

\newpage

\section{Limitations}

A key difficulty in generating data via \gls{sdscm} for a use-case like benchmarking causal inference methods is to ensure the data have meaningful structure (e.g., significant non-trivial relationships between variables). The reason for this challenge is, in part, by design: the user does not directly specify structural equations. Instead, the structural equations are determined by whatever the language model $\mP$ has already encoded. This reliance on what has been previously encoded by a pre-trained language model can sometimes be limiting, motivating a direction of future extensions of \glspl{sdscm} focused on \emph{training} \glspl{lm} to induce relationships between variables while following an input causal structure, rather than using pre-trained \glspl{lm}.

A related limitation in the current work is the need to manually account for the sensitivity of generated data to input variable phrasings. For example, in the breast cancer \glspl{sdscm}, we manually create many different phrasings of each sequence variable in order to account for this source of variability. This can be a tedious process as the number of variables grows and could be automated end-to-end in future work.

\paragraph{Risks and societal consequences.} There are many potential societal consequences of our work, which are essentially those shared by any model-agnostic application of language models. Pre-trained language models often come with inherent biases and inaccuracies. Generated data may still include such biases or inaccuracies, whether intentional or not. Any future work that builds on this work for the purposes of auditing language models will also inherit the limitations of all tools for model explainability: model explanations always have the potential to be misleading or over-simplified.

\section*{Acknowledgments}

This research was supported in part by National Science Foundation (NSF) award No. 1922658
and the Samsung Advanced Institute of Technology (under the project Next Generation Deep
Learning: From Pattern Recognition to AI). We are grateful to Roman Mutel,  Yulia Maksymiuk, and Julia Stoyanovich for involvement in early discussions of this work and feedback on data storage.

\bibliography{references}

\begin{thebibliography}{57}
\providecommand{\natexlab}[1]{#1}

\bibitem[{Angrist et~al.(1993)Angrist, Imbens, and Rubin}]{Angrist1993IdentificationOC}
Joshua~David Angrist, Guido Imbens, and Donald~B. Rubin. 1993.
\newblock Identification of causal effects using instrumental variables.

\bibitem[{Athey et~al.(2024)Athey, Imbens, Metzger, and Munro}]{athey2024wasserstein}
Susan Athey, Guido~W. Imbens, Jonas Metzger, and Evan Munro. 2024.
\newblock \href {https://doi.org/10.1016/j.jeconom.2020.09.013} {Using wasserstein generative adversarial networks for the design of monte carlo simulations}.
\newblock \emph{Journal of Econometrics}, 240(2):105076.

\bibitem[{Athey et~al.(2019)Athey, Tibshirani, and Wager}]{Athey2019Generalized}
Susan Athey, Julie Tibshirani, and Stefan Wager. 2019.
\newblock Generalized random forests.

\bibitem[{Balke and Pearl(1994)}]{Balke1994ProbabilisticEO}
Alexander Balke and Judea Pearl. 1994.
\newblock Probabilistic evaluation of counterfactual queries.
\newblock \emph{Probabilistic and Causal Inference}.

\bibitem[{Battocchi et~al.(2019)Battocchi, Dillon, Hei, Lewis, Oka, Oprescu, and Syrgkanis}]{econml}
Keith Battocchi, Eleanor Dillon, Maggie Hei, Greg Lewis, Paul Oka, Miruna Oprescu, and Vasilis Syrgkanis. 2019.
\newblock {EconML}: {A Python Package for ML-Based Heterogeneous Treatment Effects Estimation}.
\newblock https://github.com/py-why/EconML.
\newblock Version 0.x.

\bibitem[{Betti et~al.(2023)Betti, Abrate, Bonchi, and Kaltenbrunner}]{Betti2023RelevancebasedIF}
Lorenzo Betti, Carlo Abrate, Francesco Bonchi, and Andreas Kaltenbrunner. 2023.
\newblock Relevance-based infilling for natural language counterfactuals.
\newblock \emph{Proceedings of the 32nd ACM International Conference on Information and Knowledge Management}.

\bibitem[{Chatzi et~al.(2024)Chatzi, Benz, Straitouri, Tsirtsis, and Gomez-Rodriguez}]{Chatzi2024CounterfactualTG}
Ivi Chatzi, Nina~Corvelo Benz, Eleni Straitouri, Stratis Tsirtsis, and Manuel Gomez-Rodriguez. 2024.
\newblock Counterfactual token generation in large language models.
\newblock \emph{ArXiv}, abs/2409.17027.

\bibitem[{Chernozhukov et~al.(2017)Chernozhukov, Goldman, Semenova, and Taddy}]{chernozhukov2017orthogonal}
Victor Chernozhukov, Matt Goldman, Vira Semenova, and Matt Taddy. 2017.
\newblock Orthogonal machine learning for demand estimation: High dimensional causal inference in dynamic panels.
\newblock \emph{arXiv}, pages arXiv--1712.

\bibitem[{Chetverikov et~al.(2016)Chetverikov, Demirer, Duflo, Hansen, Newey, and Chernozhukov}]{chetverikov2016double}
D~Chetverikov, M~Demirer, E~Duflo, C~Hansen, WK~Newey, and V~Chernozhukov. 2016.
\newblock Double machine learning for treatment and causal parameters. 2016.

\bibitem[{Chiappa(2018)}]{Chiappa2018PathSpecificCF}
Silvia Chiappa. 2018.
\newblock Path-specific counterfactual fairness.
\newblock In \emph{AAAI Conference on Artificial Intelligence}.

\bibitem[{Chipman et~al.(2008)Chipman, George, and McCulloch}]{Chipman2008BARTBA}
Hugh~A. Chipman, Edward~I. George, and Robert~E. McCulloch. 2008.
\newblock Bart: Bayesian additive regression trees.
\newblock \emph{The Annals of Applied Statistics}, 4:266--298.

\bibitem[{Curth et~al.(2021)Curth, Svensson, Weatherall, and van~der Schaar}]{Curth2021ReallyDG}
Alicia Curth, David Svensson, James Weatherall, and Mihaela van~der Schaar. 2021.
\newblock Really doing great at estimating cate? a critical look at ml benchmarking practices in treatment effect estimation.
\newblock In \emph{NeurIPS Datasets and Benchmarks}.

\bibitem[{Curth and van~der Schaar(2021{\natexlab{a}})}]{curth2021nonparametric}
Alicia Curth and Mihaela van~der Schaar. 2021{\natexlab{a}}.
\newblock Nonparametric estimation of heterogeneous treatment effects: From theory to learning algorithms.
\newblock In \emph{Proceedings of the 24th International Conference on Artificial Intelligence and Statistics (AISTATS)}. PMLR.

\bibitem[{Curth and van~der Schaar(2021{\natexlab{b}})}]{curth2021inductive}
Alicia Curth and Mihaela van~der Schaar. 2021{\natexlab{b}}.
\newblock On inductive biases for heterogeneous treatment effect estimation.

\bibitem[{Dehejia and Wahba(1999)}]{Dehejia1999Causal}
Rajeev~H Dehejia and Sadek Wahba. 1999.
\newblock Causal effects in nonexperimental studies: Reevaluating the evaluation of training programs.
\newblock \emph{Journal of the American statistical Association}, 94(448):1053--1062.

\bibitem[{Dorie and Hill(2020)}]{Dorie2020CausalIU}
Vincent Dorie and Jennifer~L. Hill. 2020.
\newblock Causal inference using bayesian additive regression trees [r package bartcause version 1.0-4].

\bibitem[{Dubey et~al.(2024)Dubey, Jauhri, Pandey, Kadian, Al-Dahle, Letman, Mathur, Schelten, Yang, Fan et~al.}]{Dubey2024Llama}
Abhimanyu Dubey, Abhinav Jauhri, Abhinav Pandey, Abhishek Kadian, Ahmad Al-Dahle, Aiesha Letman, Akhil Mathur, Alan Schelten, Amy Yang, Angela Fan, and 1 others. 2024.
\newblock The llama 3 herd of models.
\newblock \emph{arXiv preprint arXiv:2407.21783}.

\bibitem[{Feder et~al.(2021)Feder, Keith, Manzoor, Pryzant, Sridhar, Wood-Doughty, Eisenstein, Grimmer, Reichart, Roberts, Stewart, Veitch, and Yang}]{Feder2021CausalII}
Amir Feder, Katherine~A. Keith, Emaad~A. Manzoor, Reid Pryzant, Dhanya Sridhar, Zach Wood-Doughty, Jacob Eisenstein, Justin Grimmer, Roi Reichart, Margaret~E. Roberts, Brandon~M Stewart, Victor Veitch, and Diyi Yang. 2021.
\newblock Causal inference in natural language processing: Estimation, prediction, interpretation and beyond.
\newblock \emph{Transactions of the Association for Computational Linguistics}, 10:1138--1158.

\bibitem[{Foster and Syrgkanis(2023)}]{foster2023orthogonal}
Dylan~J Foster and Vasilis Syrgkanis. 2023.
\newblock Orthogonal statistical learning.
\newblock \emph{The Annals of Statistics}, 51(3):879--908.

\bibitem[{Franklin et~al.(2014)Franklin, Schneeweiss, Polinski, and Rassen}]{franklin2014plasmode}
Jessica~M Franklin, Sebastian Schneeweiss, Jennifer~M Polinski, and Jeremy~A Rassen. 2014.
\newblock Plasmode simulation for the evaluation of pharmacoepidemiologic methods in complex healthcare databases.
\newblock \emph{Computational statistics \& data analysis}, 72:219--226.

\bibitem[{Fryer et~al.(2022)Fryer, Axelrod, Packer, Beutel, Chen, and Webster}]{Fryer2022FlexibleTG}
Zee Fryer, Vera Axelrod, Ben Packer, Alex Beutel, Jilin Chen, and Kellie Webster. 2022.
\newblock Flexible text generation for counterfactual fairness probing.
\newblock \emph{ArXiv}, abs/2206.13757.

\bibitem[{Gat et~al.(2023)Gat, Calderon, Feder, Chapanin, Sharma, and Reichart}]{Gat2023FaithfulEO}
Yair~Ori Gat, Nitay Calderon, Amir Feder, Alexander Chapanin, Amit Sharma, and Roi Reichart. 2023.
\newblock Faithful explanations of black-box nlp models using llm-generated counterfactuals.
\newblock \emph{ArXiv}, abs/2310.00603.

\bibitem[{Gentzel et~al.(2019)Gentzel, Garant, and Jensen}]{gentzel2019empirical}
Amanda Gentzel, Dan Garant, and David Jensen. 2019.
\newblock The case for evaluating causal models using interventional measures and empirical data.
\newblock In \emph{Advances in Neural Information Processing Systems}, volume~32. Curran Associates, Inc.

\bibitem[{Gentzel et~al.(2021)Gentzel, Pruthi, and Jensen}]{gentzel2021and}
Amanda~M Gentzel, Purva Pruthi, and David Jensen. 2021.
\newblock How and why to use experimental data to evaluate methods for observational causal inference.
\newblock In \emph{International Conference on Machine Learning}, pages 3660--3671. PMLR.

\bibitem[{Hao et~al.(2021)Hao, Pang, Lan, Wang, Guo, and Cheng}]{Hao2021SketchAC}
Changying Hao, Liang Pang, Yanyan Lan, Yan Wang, Jiafeng Guo, and Xueqi Cheng. 2021.
\newblock Sketch and customize: A counterfactual story generator.
\newblock \emph{ArXiv}, abs/2104.00929.

\bibitem[{Hern{\'a}n and Robins(2006)}]{Hernn2006InstrumentsFC}
Miguel~A. Hern{\'a}n and James~M. Robins. 2006.
\newblock Instruments for causal inference: An epidemiologist's dream?
\newblock \emph{Epidemiology}, 17:360--372.

\bibitem[{Hill(2011)}]{Hill2011BayesianNM}
Jennifer~L. Hill. 2011.
\newblock Bayesian nonparametric modeling for causal inference.
\newblock \emph{Journal of Computational and Graphical Statistics}, 20:217 -- 240.

\bibitem[{Holland(1985)}]{Holland1985StatisticsAC}
Paul Holland. 1985.
\newblock Statistics and causal inference.
\newblock \emph{Journal of the American Statistical Association}, 81:945--960.

\bibitem[{Im et~al.(2024)Im, Zhang, Verma, and Cho}]{Im2024UsingDA}
Daniel~Jiwoong Im, Kevin Zhang, Nakul Verma, and Kyunghyun Cho. 2024.
\newblock Using deep autoregressive models as causal inference engines.
\newblock \emph{ArXiv}, abs/2409.18581.

\bibitem[{Jin et~al.(2023{\natexlab{a}})Jin, Chen, Leeb, Gresele, Kamal, Lyu, Blin, Adauto, Kleiman-Weiner, Sachan, and Sch{\"o}lkopf}]{Jin2023CLadderAB}
Zhijing Jin, Yuen Chen, Felix Leeb, Luigi Gresele, Ojasv Kamal, Zhiheng Lyu, Kevin Blin, Fernando~Gonzalez Adauto, Max Kleiman-Weiner, Mrinmaya Sachan, and Bernhard Sch{\"o}lkopf. 2023{\natexlab{a}}.
\newblock Cladder: A benchmark to assess causal reasoning capabilities of language models.
\newblock \emph{ArXiv}, abs/2312.04350.

\bibitem[{Jin et~al.(2023{\natexlab{b}})Jin, Liu, Lyu, Poff, Sachan, Mihalcea, Diab, and Scholkopf}]{Jin2023CanLL}
Zhijing Jin, Jiarui Liu, Zhiheng Lyu, Spencer Poff, Mrinmaya Sachan, Rada Mihalcea, Mona~T. Diab, and Bernhard Scholkopf. 2023{\natexlab{b}}.
\newblock Can large language models infer causation from correlation?
\newblock \emph{ArXiv}, abs/2306.05836.

\bibitem[{Knaus et~al.(2021)Knaus, Lechner, and Strittmatter}]{knaus2021machine}
Michael~C Knaus, Michael Lechner, and Anthony Strittmatter. 2021.
\newblock Machine learning estimation of heterogeneous causal effects: Empirical monte carlo evidence.
\newblock \emph{The Econometrics Journal}, 24(1):134--161.

\bibitem[{Kr{\"a}mer et~al.(2013)Kr{\"a}mer, Green, Pollard, and Tugendreich}]{Krmer2013CausalAA}
Andreas Kr{\"a}mer, Jeff Green, Jack Pollard, and Stuart Tugendreich. 2013.
\newblock Causal analysis approaches in ingenuity pathway analysis.
\newblock \emph{Bioinformatics}, 30:523 -- 530.

\bibitem[{K{\"u}nzel et~al.(2019)K{\"u}nzel, Sekhon, Bickel, and Yu}]{kunzel2019metalearners}
S{\"o}ren~R K{\"u}nzel, Jasjeet~S Sekhon, Peter~J Bickel, and Bin Yu. 2019.
\newblock Metalearners for estimating heterogeneous treatment effects using machine learning.
\newblock \emph{Proceedings of the national academy of sciences}, 116(10):4156--4165.

\bibitem[{Kusner et~al.(2017)Kusner, Loftus, Russell, and Silva}]{Kusner2017CounterfactualF}
Matt~J. Kusner, Joshua~R. Loftus, Chris Russell, and Ricardo Silva. 2017.
\newblock Counterfactual fairness.
\newblock \emph{ArXiv}, abs/1703.06856.

\bibitem[{Kıcıman et~al.(2023)Kıcıman, Ness, Sharma, and Tan}]{Kcman2023CausalRA}
Emre Kıcıman, Robert~Osazuwa Ness, Amit Sharma, and Chenhao Tan. 2023.
\newblock Causal reasoning and large language models: Opening a new frontier for causality.
\newblock \emph{ArXiv}, abs/2305.00050.

\bibitem[{Lei and Cand{\`e}s(2020)}]{Lei2020ConformalIO}
Lihua Lei and Emmanuel~J. Cand{\`e}s. 2020.
\newblock Conformal inference of counterfactuals and individual treatment effects.
\newblock \emph{Journal of the Royal Statistical Society: Series B (Statistical Methodology)}, 83.

\bibitem[{Li et~al.(2023)Li, Xu, Miao, Zhou, and Qian}]{Li2023PromptingLL}
Yongqi Li, Mayi Xu, Xin Miao, Shen Zhou, and Tieyun Qian. 2023.
\newblock Prompting large language models for counterfactual generation: An empirical study.
\newblock In \emph{International Conference on Language Resources and Evaluation}.

\bibitem[{Liu et~al.(2024)Liu, Xu, Wu, Yuan, Yang, Zhou, Liu, Guan, Wang, Yu, McAuley, Ai, and Huang}]{Liu2024LargeLM}
Xiaoyu Liu, Paiheng Xu, Junda Wu, Jiaxin Yuan, Yifan Yang, Yuhang Zhou, Fuxiao Liu, Tianrui Guan, Haoliang Wang, Tong Yu, Julian McAuley, Wei Ai, and Furong Huang. 2024.
\newblock Large language models and causal inference in collaboration: A comprehensive survey.
\newblock \emph{ArXiv}, abs/2403.09606.

\bibitem[{Louizos et~al.(2017)Louizos, Shalit, Mooij, Sontag, Zemel, and Welling}]{Louizos2017CausalEI}
Christos Louizos, Uri Shalit, Joris~M. Mooij, David~A. Sontag, Richard~S. Zemel, and Max Welling. 2017.
\newblock Causal effect inference with deep latent-variable models.
\newblock In \emph{Neural Information Processing Systems}.

\bibitem[{Mackey et~al.(2018)Mackey, Syrgkanis, and Zadik}]{mackey2018orthogonal}
Lester Mackey, Vasilis Syrgkanis, and Ilias Zadik. 2018.
\newblock Orthogonal machine learning: Power and limitations.
\newblock In \emph{International Conference on Machine Learning}, pages 3375--3383. PMLR.

\bibitem[{McDuff et~al.(2022)McDuff, Song, Lee, Vineet, Vemprala, Gyde, Salman, Ma, Sohn, and Kapoor}]{mcduff2022causalcity}
Daniel McDuff, Yale Song, Jiyoung Lee, Vibhav Vineet, Sai Vemprala, Nicholas~Alexander Gyde, Hadi Salman, Shuang Ma, Kwanghoon Sohn, and Ashish Kapoor. 2022.
\newblock Causalcity: Complex simulations with agency for causal discovery and reasoning.
\newblock In \emph{Conference on Causal Learning and Reasoning}, pages 559--575. PMLR.

\bibitem[{Neal et~al.(2020)Neal, Huang, and Raghupathi}]{Neal2020RealCauseRC}
Brady Neal, Chin-Wei Huang, and Sunand Raghupathi. 2020.
\newblock Realcause: Realistic causal inference benchmarking.
\newblock \emph{ArXiv}, abs/2011.15007.

\bibitem[{Nie and Wager(2021)}]{nie2021quasi}
Xinkun Nie and Stefan Wager. 2021.
\newblock Quasi-oracle estimation of heterogeneous treatment effects.
\newblock \emph{Biometrika}, 108(2):299--319.

\bibitem[{Pawlowski et~al.(2020)Pawlowski, Coelho~de Castro, and Glocker}]{pawlowski2020deep}
Nick Pawlowski, Daniel Coelho~de Castro, and Ben Glocker. 2020.
\newblock Deep structural causal models for tractable counterfactual inference.
\newblock \emph{Advances in neural information processing systems}, 33:857--869.

\bibitem[{Peters et~al.(2017)Peters, Janzing, and Sch{\"o}lkopf}]{Peters2017ElementsOC}
Jonas Peters, Dominik Janzing, and Bernhard Sch{\"o}lkopf. 2017.
\newblock Elements of causal inference: Foundations and learning algorithms.

\bibitem[{Radford et~al.(2019)Radford, Wu, Child, Luan, Amodei, and Sutskever}]{Radford2019LanguageMA}
Alec Radford, Jeff Wu, Rewon Child, David Luan, Dario Amodei, and Ilya Sutskever. 2019.
\newblock Language models are unsupervised multitask learners.

\bibitem[{Ravfogel et~al.(2024)Ravfogel, Svete, Snæbjarnarson, and Cotterell}]{ravfogel2024gumbelcounterfactualgenerationlanguage}
Shauli Ravfogel, Anej Svete, Vésteinn Snæbjarnarson, and Ryan Cotterell. 2024.
\newblock \href {https://arxiv.org/abs/2411.07180} {Gumbel counterfactual generation from language models}.
\newblock \emph{Preprint}, arXiv:2411.07180.

\bibitem[{Sanchez and Tsaftaris(2022)}]{Sanchez2022DiffusionCM}
Pedro Sanchez and Sotirios~A. Tsaftaris. 2022.
\newblock Diffusion causal models for counterfactual estimation.
\newblock In \emph{CLEaR}.

\bibitem[{Schuler et~al.(2017)Schuler, Jung, Tibshirani, Hastie, and Shah}]{schuler2017synth}
Alejandro Schuler, Ken Jung, Robert Tibshirani, Trevor Hastie, and Nigam Shah. 2017.
\newblock Synth-validation: Selecting the best causal inference method for a given dataset.
\newblock \emph{arXiv preprint arXiv:1711.00083}.

\bibitem[{Shalit et~al.(2017)Shalit, Johansson, and Sontag}]{shalit2017estimating}
Uri Shalit, Fredrik~D Johansson, and David Sontag. 2017.
\newblock Estimating individual treatment effect: generalization bounds and algorithms.
\newblock In \emph{International conference on machine learning}, pages 3076--3085. PMLR.

\bibitem[{Vashishtha et~al.(2023)Vashishtha, Reddy, Kumar, Bachu, Balasubramanian, and Sharma}]{Vashishtha2023CausalIU}
Aniket Vashishtha, Abbavaram~Gowtham Reddy, Abhinav Kumar, Saketh Bachu, Vineeth~N. Balasubramanian, and Amit Sharma. 2023.
\newblock Causal inference using llm-guided discovery.
\newblock \emph{ArXiv}, abs/2310.15117.

\bibitem[{Vegetabile(2021)}]{Vegetabile2021OnTD}
Brian~G. Vegetabile. 2021.
\newblock On the distinction between "conditional average treatment effects" (cate) and "individual treatment effects" (ite) under ignorability assumptions.
\newblock \emph{ArXiv}, abs/2108.04939.

\bibitem[{Wager and Athey(2018)}]{Wager2018Estimation}
Stefan Wager and Susan Athey. 2018.
\newblock Estimation and inference of heterogeneous treatment effects using random forests.
\newblock \emph{Journal of the American Statistical Association}, 113(523):1228--1242.

\bibitem[{Wang et~al.(2024)Wang, Qiu, Yue, Guo, Zeng, Feng, and Shen}]{Wang2024ASO}
Yongjie Wang, Xiaoqi Qiu, Yu~Yue, Xu~Guo, Zhiwei Zeng, Yuhong Feng, and Zhiqi Shen. 2024.
\newblock A survey on natural language counterfactual generation.
\newblock \emph{ArXiv}, abs/2407.03993.

\bibitem[{Wendling et~al.(2018)Wendling, Jung, Callahan, Schuler, Shah, and Gallego}]{wendling2018comparing}
Thierry Wendling, Kenneth Jung, Alison Callahan, Alejandro Schuler, Nigam~H Shah, and Blanca Gallego. 2018.
\newblock Comparing methods for estimation of heterogeneous treatment effects using observational data from health care databases.
\newblock \emph{Statistics in medicine}, 37(23):3309--3324.

\bibitem[{Zecevic et~al.(2023)Zecevic, Willig, Dhami, and Kersting}]{Zecevic2023CausalPL}
M.~Zecevic, Moritz Willig, Devendra~Singh Dhami, and Kristian Kersting. 2023.
\newblock Causal parrots: Large language models may talk causality but are not causal.
\newblock \emph{ArXiv}, abs/2308.13067.

\end{thebibliography}

\appendix

\section{Formal definition of sequence-driven structural causal models}\label{sec:full_definitions}

We first introduce notation preliminaries in \Cref{sec:notation} before formally defining our procedure in \Cref{sec:sdscm}.

\subsection{Preliminaries}\label{sec:notation}
Let lowercase letter with tilde $\tv$ denote a random variable, where $\tv=v$ denotes the value it obtains. Let boldface capital letter $\bV = \{\tv_1, \ldots, \tv_n\}$ denote a set of variables with value $\bV = \bv$, capital $P_{\tv}$ denote the cumulative distribution function of $\tv$, and lowercase $p_{\tv}$ denote the density (or mass) function. Let $P_{\tv \mid \tx=x}$ denote the conditional distribution of $\tv$ given $\tx=x$ and $P_{\tv \mid \tx}$ denote the collection of $P_{\tv \mid \tx=x}$ for all $x$. A sequence, or string, is an ordered collection of tokens. We represent this either as a tuple (e.g., sequence $v=(w_1, \ldots, w_T)$ has tokens $w_t$), or interchangeably as a single string (e.g., $v = w_{1:T} \equiv \bigoplus_{t=1}^{T} w_t$, where $\oplus$ represents string concatenation).

\begin{definition}[Language model]
    Given a vocabulary $\mathbb{V}$ of possible tokens, we define a language model $\mP$ as a joint distribution over any sequence of tokens $v = (w_1, \ldots, w_T) \in \bigtimes_{t=1}^T \mathbb{V}$, where $\mP(v) = \prod_{t = 1}^{T} \mP(w_t \mid w_{1:(t-1)})$.
\end{definition}

\begin{definition}[Structural causal model]
    We define a \acrfull{scm} as a 4-tuple $\fC = (\bV, \bU, \bF, P_\bU)$. In this tuple, $\bV$ is a set of observed variables, $\bU$ a set of unobserved (exogenous) variables, $\bF$ a set of functions $\{f_i\}_{i=1}^{|\bV|}$ for each $\tv_i \in \bV$ such that $\tv_i = f_i(\bPA_i, \bU_i)$ where $\bPA_i \subseteq \bV \setminus \{ \tv_i\}$ represents the causal parents of $\tv_i$ and $\bU_i \subseteq \bU$, and $P_\bU$ a distribution over $\bU$. A causal model can be represented visually as a \acrfull{dag} with nodes for $\bU, \bV$ and directed edges for $\bF$. \glspl{scm} entail an observational distribution $P^{\fC}$ across variables $\bV \cup \bU$.
\end{definition}

\begin{definition}[Interventional distribution]
    An \gls{scm} $\fC$ also entails the distribution of any subset of variables in $\bV \cup \bU$ following atomic intervention $I = \text{do}\left(\tv_i := v\right)$, which replaces the structural mechanism $f_i$ with fixed value $v$. Interventions can also be extended to general modifications of $f_i$. We denote an \gls{scm} after intervention $I$ as $\fC^{\doI}$ and the resulting distribution as $P^{\fC; \doI}$.\footnote{Our notational conventions for interventional and counterfactual distributions follow \cite{Peters2017ElementsOC}.}
\end{definition}

Counterfactual distributions are computed in a similar fashion, but first conditioning $P_{\bU}$ on a particular context before performing an intervention. Where ambiguous, we use an asterisk to denote counterfactual versions $\bV^*$ of factual variables $\bV$ \cite{Balke1994ProbabilisticEO}. 

\begin{definition}[Counterfactual distribution]
     Counterfactual variable $\bY^*$ given a factual observation $\bz$ and intervention $\text{do}(I)$ (where $\bY, \bZ \subseteq \bV$) can be computed via a three-step procedure often referred to as `abduction, action, prediction.' Abduction uses observed evidence to obtain $P_{\bU \mid \bZ=\bz}$ from $P_{\bU}$. Action performs intervention $\text{do}(I)$ to obtain modified \gls{scm} $\fC^{\doI}$. Prediction computes the probability of $\bY^*$ from $\fC^{\doI}$ and $P_{\bU \mid \bZ=\bz}$. For general intervention $I$ and observed assignment $\bZ = \bz$, we denote the counterfactual distribution $P^{\fC \mid \bZ = \bz; \doI}$.
\end{definition}

\subsection{\Glsentrylongpl{sdscm}}\label{sec:sdscm}

Consider a collection of ordered random variables $(\tv_1, \tv_2, \tv_3, \ldots)$, whose sample spaces $\Omega_{\tv_i}$ each consist of sets of sequences. We define $\tv_{1:m} \equiv \bigoplus_{t=1}^{m} \tv_t$ as the concatenation of the sequences themselves. The sample space for the concatenation of sequences is the cartesian product of the constituent sample spaces $\bigtimes_{t=1}^{m} \Omega_{\tv_t}$. For brevity, we will use the term \textbf{sequence variable} to refer to a random variable whose sample space is a set of

\noindent \begin{minipage}{0.48\textwidth}
\vspace{-1em}
\begin{algorithm}[H]
\setstretch{1.1}
\caption{A single \gls{sdscm} sample from the observational distribution}\label[algorithm]{alg:observational_sampling}
\KwInput{$\fB = (\bV, \bU, \mG, \mP(\cdot), \tau)$}
\KwReturns{$\mathbf{s} = \left(u_1, \ldots, u_{|\bU|}, v_1, \ldots, v_{|\bV|}\right)$}
$\mathbf{s} \gets ()$

\For{$\tx_t \in \tau$}{
    $\textrm{PA}_\tau \gets \{t' : \tx_{t'} \in \textrm{PA}_{\tx_t}\}$ ordered by $\tau$
    
    $x_{\textrm{PA}_\tau} \gets \bigoplus_{x \in \mathbf{s}\left[\textrm{PA}_\tau \right]} x$
    
    $\mathbf{p}_{\tx_t} \gets []$
    
    \For{$k \in 1, \ldots, \left|\Omega_{\tx_t}\right|$}{
        $x \gets \Omega_{\tx_t}[k]$
        
        $\mathbf{p}_{\tx_t}[k] \gets \mP\left( x_{\textrm{PA}_\tau} \oplus x\right)$
    }
    $P_{\textrm{tot}} \gets \sum_k \mathbf{p}_{\tx_t}[k]$
    
    $j \sim \textrm{Multinomial}(\mathbf{p}_{\tx_t} / P_{\textrm{tot}})$
    
    $x_t \gets \Omega_{\tx_t}[j]$
    
    $\textrm{append}(\mathbf{s}, x_t)$
}
\textbf{return} $\mathbf{s}$
\end{algorithm}
\end{minipage}\\

\noindent sequences. Two straightforward abstractions allow us to define \glspl{sdscm}: domain-restricted sampling and parent-only concatenation.

\begin{definition}[Domain-restricted sampling]
    Given language model $\mP$, some prior inputs $C$, and a sequence variable $\tv_i$ with sample space $\Omega_{\tv_i}$, domain-restricted sampling defines a distribution $\mP_{\tv_i \mid C}$ over sample space $\Omega_{\tv_i}$ simply by tabulating and subsequently normalizing the output probabilities for each possible $v \in \Omega_{\tv_i}$ conditional on prior inputs $C$:
    $\mP_{\tv_i \mid C}(v) \equiv \frac{\mP(v \mid C)}{\sum_{v' \in \Omega_{\tv_i}} \mP(v' \mid C)}.$
\end{definition}

\begin{definition}[Parent-only concatenation]
    Given \gls{dag} $\mG$ over $m$ sequence variables $(\tv_1, \ldots, \tv_m)$ and a topological ordering $\tau$ consistent with $\mG$, parent-only concatenation defines $\left(\tv_i \mid \bPA_i \right) \equiv \left(\bigoplus_{t \in \bPA_i} \tv_t \right) \oplus \tv_i$, where $\bPA_i$ are the parents of $\tv_i$ in $\mG$ ordered according to $\tau$.
\end{definition}

\noindent Given a \gls{dag} $\mG$ and a language model $\mP$, a corresponding sequence-driven \gls{scm} defines a sample space of sequences for each variable in $\mG$ and provides access to observational, interventional, and counterfactual distributions as follows.

\begin{definition}[\Glsentryfull{sdscm}]
    We define a \acrlong{sdscm} as a 5-tuple $\fB = (\bV, \bU, \mG, \mP, \tau)$, where
    \begin{itemize}[noitemsep]
        \item $\bV$ is a set of finite-domain endogenous/observed sequence variables and $\bU$ a set of finite-domain exogenous/unobserved sequence variables;
        \item $\mG$ is a \gls{dag} over the variables $\tx_i$ in $\bV \cup \bU$ where $\bPA_i \subseteq \left(\bV \cup \bU \right) \setminus \{\tx_i\}$;
        \item $\mP$ is a language model trained on prior inputs $C$ whose vocabulary $\mathbb{V}$ contains all tokens used in $\Omega_{\bV}, \Omega_{\bU}$; and
        \item $\tau$ is an arbitrary fixed topological ordering of $\bV \cup \bU$ consistent with $\mG$.
    \end{itemize}
    An \gls{sdscm} uses $\mP$ to define an \textbf{observational distribution} over the variables in $\bV \cup \bU$ that factorizes according to $\mG$ via domain-restricted ancestral sampling and parent-only concatenation with $\tau$: $P^{\fB} \equiv \prod_{\tx_t \in \tau} \mP_{\tx_t \mid C, \bPA_t}$. This procedure is shown in \Cref{alg:observational_sampling}.
\end{definition}

The key difference between an \gls{scm} and an \gls{sdscm} is that all variables have at least one common ancestor --- the prior inputs $C$ that were used to train the language model, if any. It is however possible to train the \gls{lm} to induce distributions over the desired variables given this setup. As with the observational distribution, domain-restricted ancestral sampling and parent-only concatenation also allow us to define interventional and counterfactual distributions.

\begin{definition}[Sequence-driven interventional distribution]\label[definition]{def:sdscm_interventional_distribution}
An \gls{sdscm} $\fB$ entails the distribution of any subset of variables in $\bV \cup \bU$ following intervention $I = \text{do}\left(\tv_i = v\right)$ by replacing variable $\tv_i$ with value $v$, and otherwise sampling in the same manner. As with an \gls{scm}, we denote an \gls{sdscm} after intervention $I$ as $\fB^{\doI}$ and the resulting interventional distribution as $P^{\fB; \doI}$. This is computed for intervention $\doVi$ as follows: $P^{\fB; \doVi} \equiv \prod_{\tx_t \in \tau} \mP_{\tx_t \mid C, \tv_i=v, \bPA'_t}$, 
where $\bPA'_t = \bPA_t \setminus \{\tv_i\}$. This procedure is shown in \Cref{alg:interventional_sampling}.
\end{definition}

In order to admit unique answers to counterfactual queries, we define abduction for an \gls{sdscm} given evidence $\bZ=\bz$ as the setting of values $\bU=\bu$ and any evidence upstream of the intervention, rather than a distribution $P_{\bU \mid \bZ=\bz}$.\footnote{Other choices can be explored here, which we leave to future extensions of \glspl{sdscm}.} In order to obtain such values $\bu$, one needs access to more than just the observed data and language model $\mP$ --- obtaining $\bu$ requires performing bookkeeping \emph{during the data generation process}. This is a restatement of the fact that computing point counterfactuals in \glspl{scm} requires causal mechanisms that are invertible with respect to the noise variables in order to uniquely reconstruct the noise

\noindent \begin{minipage}{0.48\textwidth}
\vspace{-1em}
\begin{algorithm}[H]
\setstretch{1.15}
\caption{A single \gls{sdscm} sample from the interventional distribution for $\doVi$}\label[algorithm]{alg:interventional_sampling}
\KwInput{$\doVi, \fB = (\bV, \bU, \mG, \mP(\cdot), \tau)$}
\KwReturns{$\mathbf{s} = \left(u_1, \ldots, u_{|\bU|}, v_1, \ldots, v_{|\bV|}\right)$}
$\mathbf{s} \gets ()$

\For{$\tx_t \in \tau$}{
    \If{$\tx_t \equiv \tv_i$}{
        $x_t \gets v$
    }
    \Else{
        $\textrm{PA}_\tau \gets \{t' : \tx_{t'} \in \textrm{PA}_{\tx_t}\}$ ordered by $\tau$
        
        $x_{\textrm{PA}_\tau} \gets \bigoplus_{x \in \mathbf{s}\left[\textrm{PA}_\tau \right]} x$
        
        $\mathbf{p}_{\tx_t} \gets []$
        
        \For{$k \in 1, \ldots, \left|\Omega_{\tx_t}\right|$}{
            $x \gets \Omega_{\tx_t}[k]$
            
            $\mathbf{p}_{\tx_t}[k] \gets \mP\left( x_{\textrm{PA}_\tau} \oplus x\right)$
        }
        $P_{\textrm{tot}} \gets \sum_k \mathbf{p}_{\tx_t}[k]$
        
        $j \sim \textrm{Multinomial}(\mathbf{p}_{\tx_t} / P_{\textrm{tot}})$
        
        $x_t \gets \Omega_{\tx_t}[j]$
    }
    $\textrm{append}(\mathbf{s}, x_t)$
}
\textbf{return} $\mathbf{s}$
\end{algorithm}
\end{minipage}\\

\noindent that produced a given observation. Because our primary application of \glspl{sdscm} in this work is data \emph{generation}, such bookkeeping is possible in all our use cases.

\begin{definition}[Sequence-driven counterfactual distribution]\label[definition]{def:sdscm_counterfactual_distribution}
Counterfactual sequence variable $\bY^*$ given factual evidence $\bz$ and intervention $\text{do}(\bX = \bx)$ (where $\bX, \bY, \bZ \subseteq \bV$) can be computed for an \gls{sdscm} $\fB$ whenever the exogenous setting $\bu=\{u_1, u_2, \ldots, u_{|\bU|}\}$ that generated evidence $\bz$ is known. As with an \gls{scm}, for intervention $I$ and observed $\bZ = \bz$, we denote the counterfactual distribution $P^{\fB \mid \bZ = \bz; \doI}$. This is computed for evidence $\bz$, exogenous conditions $\bu$, and intervention $\doVi$ as follows: $P^{\fB \mid \bZ = \bz; \doVi} \equiv \prod_{\tx_t \in \tau} \mP_{\tx_t \mid C, \bU=\bu, \bZ' = \bz', \tv_i=v, \bPA''_t}$,
where $\bZ' \subseteq \bZ$ contains all non-descendants of $\tv_i$ present in $\bZ$, and $\bPA''_t = \bPA_t \setminus \left( \bU \cup \bZ' \cup \{\tv_i\} \right)$. This procedure is shown in \Cref{alg:counterfactual_sampling_appendix}.
\end{definition}

In plain terms, counterfactual sampling sets not only an intervention $\doVi$ but also exogenous variables $\bU=\bu$ and upstream evidence in order to sample a hypothetical alternative that corresponds to the particular unit in question.

In summary, data can be generated from an 

\noindent \begin{minipage}{0.48\textwidth}
\vspace{-1em}
\begin{algorithm}[H]
\setstretch{1.1}
\caption{A single \gls{sdscm} sample from the counterfactual distribution given observation $\mathbf{s}_{\textrm{obs}}$}\label[algorithm]{alg:counterfactual_sampling_appendix}
\KwInput{$\mathbf{s}_{\textrm{obs}} = \left(u_1, \ldots, u_{|\bU|}, v_1, \ldots, v_{|\bV|}\right) \newline \doVi, \fB = (\bV, \bU, \mG, \mP(\cdot), \tau)$}
\KwReturns{$\mathbf{s}^* = \left(u_1, \ldots, u_{|\bU|}, v^*_1, \ldots, v^*_{|\bV|}\right)$}
$\mathbf{s}^* \gets (u_1, \ldots, u_{|\bU|})$

$\textrm{ND}_i \gets$ non-descendants of $\tv_i$ in $\mG$

\For{$\tx_t \in \tau \setminus \bU$}{
    \If{$\tx_t \equiv \tv_i$}{
        $x_t \gets v$
    }
    \ElseIf{$\tx_t \in \textrm{ND}_i$}{
        $x_t \gets \mathbf{s}_{\textrm{obs}}[t]$
    }
    \Else{
        $\textrm{PA}_\tau \gets \{t' : \tx_{t'} \in \textrm{PA}_{\tx_t}\}$ ordered by $\tau$
        
        $x_{\textrm{PA}_\tau} \gets \bigoplus_{x \in \mathbf{s}^*\left[\textrm{PA}_\tau \right]} x$
        
        $\mathbf{p}_{\tx_t} \gets []$
        
        \For{$k \in 1, \ldots, \left|\Omega_{\tx_t}\right|$}{
            $x \gets \Omega_{\tx_t}[k]$
            
            $\mathbf{p}_{\tx_t}[k] \gets \mP\left( x_{\textrm{PA}_\tau} \oplus x\right)$
        }
        $P_{\textrm{tot}} \gets \sum_k \mathbf{p}_{\tx_t}[k]$
        
        $j \sim \textrm{Multinomial}(\mathbf{p}_{\tx_t} / P_{\textrm{tot}})$
        
        $x_t \gets \Omega_{\tx_t}[j]$
    }
    $\textrm{append}(\mathbf{s}^*, x_t)$
}
\textbf{return} $\mathbf{s}^*$
\end{algorithm}
\end{minipage}\\

\noindent\gls{sdscm} by domain-restricted forward sampling variables in topological order, and, with adequate bookkeeping, both interventional and counterfactual samples can also be drawn. The key difficulty in generating data this way that is also useful for benchmarking causal inference methods is to ensure it has meaningful structure. In short, it is easy to generate data, but more difficult to generate \emph{useful} data. The reason for this challenge is that we do not directly specify the structural equations; rather, \emph{the structural equations are determined by whatever the language model $\mP$ has already encoded}.

\section{Full description of the breast cancer \glsentryshortpl{sdscm}}\label{appendix:full_bcancer_description}

\begin{figure*}[h!]
    \centering
    \begin{minipage}{\textwidth}
        \centering
        \includegraphics[width=\textwidth]{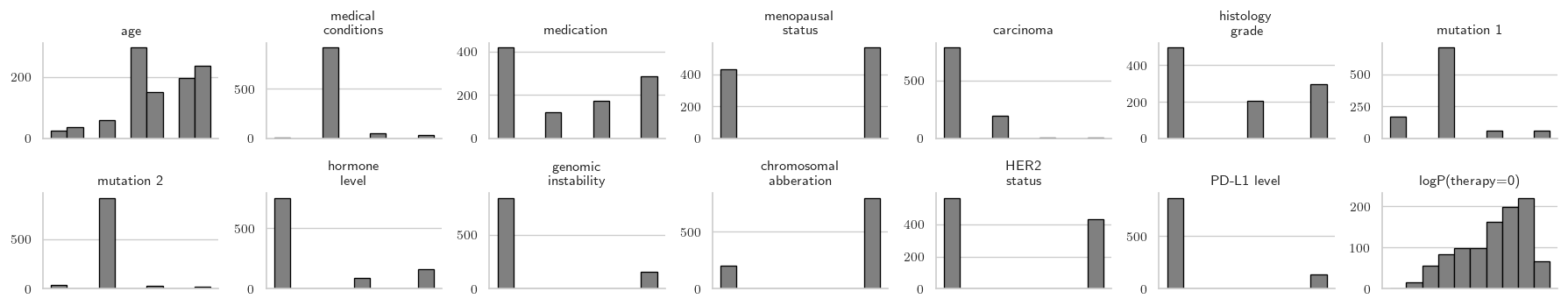}\\
        \customlabel{fig:bcancer_hist}{a}{\ref{fig:bcancer_dataset_plots}}
    \end{minipage}
    \begin{minipage}{0.32\textwidth}
        \centering
        \includegraphics[width=\textwidth]{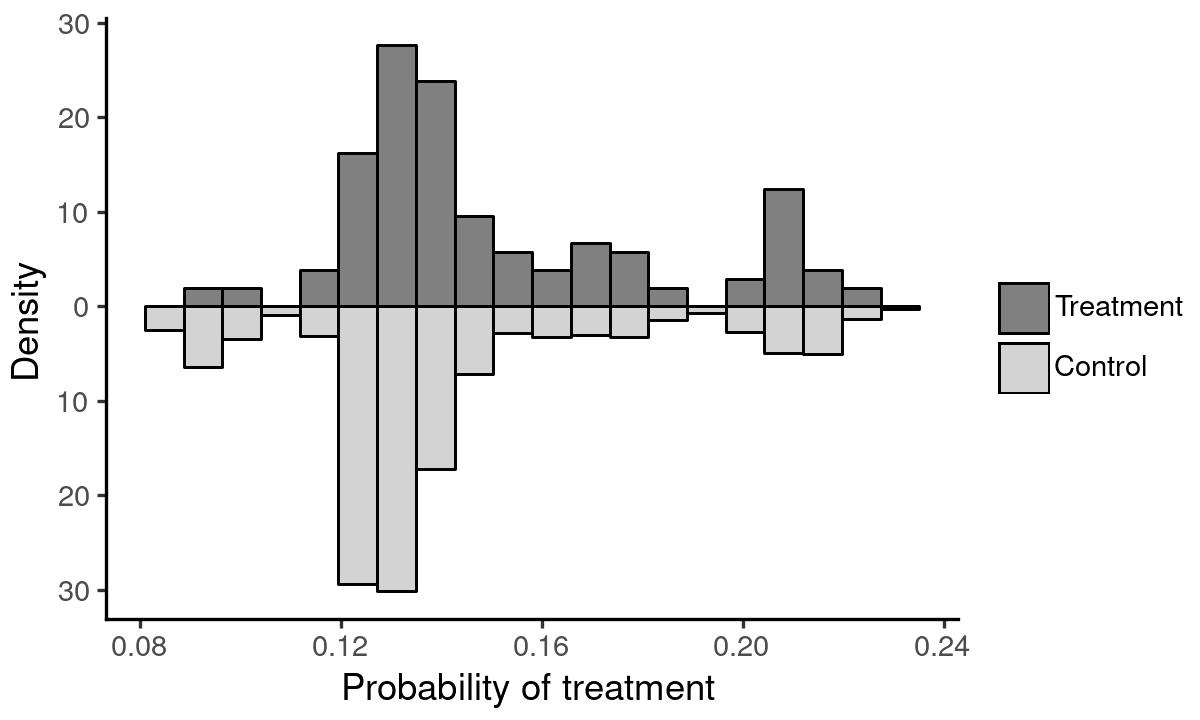}\\
        \customlabel{fig:bcancer_propensity}{b}{\ref{fig:bcancer_dataset_plots}}
    \end{minipage}
    \begin{minipage}{0.33\textwidth}
        \centering
        \includegraphics[width=\textwidth]{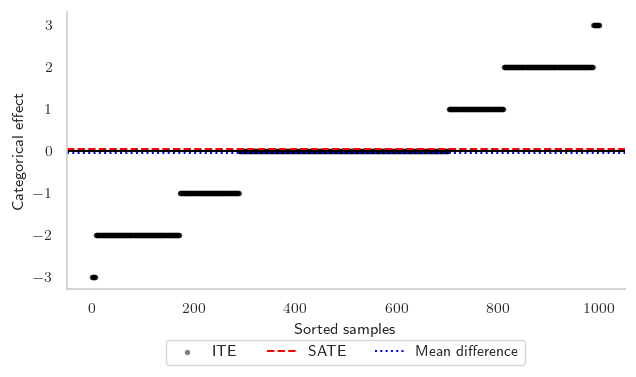}\\
        \customlabel{fig:bcancer_ites_binary}{c}{\ref{fig:bcancer_dataset_plots}}
    \end{minipage}
    \begin{minipage}{0.33\textwidth}
        \centering
        \includegraphics[width=\textwidth]{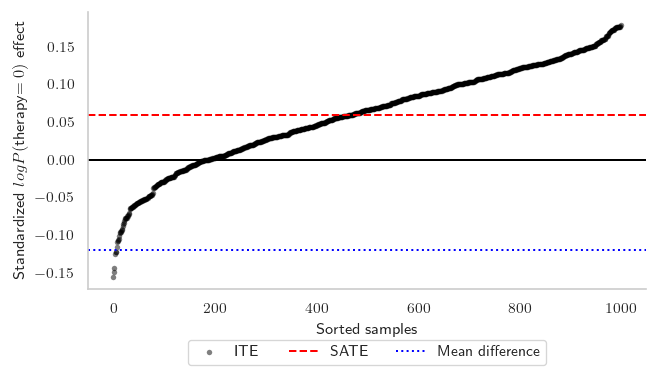}\\
        \customlabel{fig:bcancer_ites}{d}{\ref{fig:bcancer_dataset_plots}}
    \end{minipage}
    \caption{An example dataset generated by the breast cancer \gls{sdscm} using Llama-3-8b, showing \textbf{(a)} features, \textbf{(b)} propensity scores, \textbf{(c)} categorical outcome \glspl{ite}, and \textbf{(d)} continuous outcome \glspl{ite}.} 
    \label{fig:bcancer_dataset_plots}
\end{figure*}

The 14 covariates in the breast cancer \glspl{sdscm} are defined generally below. For each covariate, 10 different phrasings are considered, resulting in a sample space of $10^{14}$ possible sequences. For example, for the covariate $\tu_1$ that represents `age,' with $\Omega_{\tu_1}=(25, 35, 45, 55, 65, 75, 85)$, two possible phrasings are:
\begin{enumerate}[noitemsep]
    \item \emph{A $\tu_1$-year-old woman seeks consultation at the oncology clinic after being recently diagnosed with invasive breast cancer.}
    \item \emph{At the oncology clinic, a $\tu_1$-year-old woman is evaluated following a recent diagnosis of invasive breast carcinoma.}
\end{enumerate}
We consider 50 different variations of this \gls{sdscm}, where the sample space for a given \gls{sdscm} is defined by choosing a randomly sampled phrasing from among the possible phrasings for each of the covariates. For each of the 50 \glspl{sdscm}, 20 datasets (each of size 1000) are sampled. Each covariate and corresponding (ordered) sample space is defined as follows.
\begin{enumerate}[noitemsep]
    \item $\tu_1$: age, $\Omega_{\tu_1}=$ (25, 35, 45, 55, 65, 75, 85)
    \item $\tu_2$: medical condition, $\Omega_{\tu_2}=$ (hypertension, type 2 diabetes mellitus, hyperlipidemia, osteoporosis)
    \item $\tu_3$: medications, $\Omega_{\tu_3}=$ (lisinopril, metformin, atorvastatin, calcium carbonate)
    \item $\tu_4$: menopausal status, $\Omega_{\tu_4}=$ (pre-menopausal, post-menopausal)
    \item $\tx_1$: type of carcinoma, $\Omega_{\tx_1}=$ (invasive ductal carcinoma (IDC), invasive lobular carcinoma, medullary carcinoma, tubular carcinoma)
    \item $\tx_2$: histology grade, $\Omega_{\tx_2}=$ (grade 1, grade 2, grade 3)
    \item $\tx_3$: genetic mutation 1, $\Omega_{\tx_3}=$ (TP53, PIK3CA, BRCA1, BRCA2)
    \item $\tx_4$: genetic mutation 2, $\Omega_{\tx_4}=$ (TP53, PIK3CA, BRCA1, BRCA2)
    \item $\tx_5$: level of hormone receptor expression, $\Omega_{\tx_5}=$ (low, moderate, high)
    \item $\tx_6$: genomic instability score, $\Omega_{\tx_6}=$ (low, high)
    \item $\tx_7$: chromosomal aberration strength, $\Omega_{\tx_7}=$ (significant, minor)
    \item $\tx_8$: HER2 status, $\Omega_{\tx_8}=$ (positive, negative)
    \item $\tt$: PD-L1 expression levels, $\Omega_{\tt}=$ (low, high)
    \item $\ty$: therapy plan, $\Omega_{\ty}=$ (initiate an aromatase inhibitor therapy, administer a combination of a PARP inhibitor and chemotherapy, start a regimen of trastuzumab and pertuzumab, begin treatment with a checkpoint inhibitor such as pembrolizumab)
\end{enumerate}
The following is an example sequence randomly sampled from one possible choice of phrasings:
\begin{quote}
    \tiny
    A 65-year-old woman comes to the oncology department with a recent diagnosis of invasive breast carcinoma. Her prior medical history includes hyperlipidemia. This has been managed with lisinopril. This post-menopausal woman has no prior history of breast surgeries or hormone replacement therapy. Following a detailed assessment with imaging and biopsy, the results from the biopsy were analyzed and disclosed the following. The pathology report indicates the tumor is tubular carcinoma. The tumor's histology is rated as grade 3. The tumor shows an elevated mutation burden, with particular mutations detected in the BRCA2 gene in addition to the TP53 gene. The immunohistochemistry results display robust positive staining for estrogen receptor (ER) and progesterone receptor (PR), indicating high levels of expression. The level of genomic instability in the tumor is described as low. This implies that chromosomal aberrations are minor. Immunohistochemistry reveals HER2 as negative while FISH confirms that HER2 amplification is not present. Programmed death-ligand 1 (PD-L1) expression in the tumor is low with no distant metastases found in the imaging studies. Considering the comprehensive findings and the patient's health and treatment history, which treatment strategies are most suitable for this patient? The best option is to begin treatment with a checkpoint inhibitor such as pembrolizumab.
\end{quote}
With each sample space indexed according to the order of their values above (with indexes starting at zero), the above text sequence corresponds to the observation 
\begin{align*}
(\tu_1, \ldots, \tu_4, \tx_1, \ldots, \tx_8, \tt, \ty) =\\
(4, 2, 0, 1, 3, 2, 3, 0, 2, 0, 1, 1, 0, 3).
\end{align*}
The full set of possible examples and code to generate this \gls{sdscm} and corresponding data (in our case generated using V100 and RTX8000 GPUs) is available in our repository at \url{https://github.com/lbynum/sequence-driven-scms}.

\begin{table*}
    \footnotesize
    \caption{\gls{sate} prediction results for methods that directly target \glspl{ate}.}
    \label{tab:bcancer_ate_results}
    \centering
    \begin{tabular}{lllllllllll}
        \\\toprule
        \multirow{3}{*}{Method} & \multicolumn{4}{c}{$\mP=$ GPT-2} & \multicolumn{4}{c}{$\mP=$ Llama-3-8b} \\ 
        & \multicolumn{2}{c}{$R^2$} & \multicolumn{2}{c}{RMSE} & \multicolumn{2}{c}{$R^2$} & \multicolumn{2}{c}{RMSE} \\
        & All Cov. & Hidden & All Cov. & Hidden & All Cov. & Hidden & All Cov. & Hidden \\
        \midrule
        T-Only OLS     & 0.6047 & 0.6047 & 0.0172  & 0.0172  & 0.5082 & 0.5082 & 0.0091 & 0.0091 \\\midrule
        BART & \textbf{0.9999} & \textbf{0.8794} & \textbf{0.0003} & \textbf{0.0095} & \textbf{0.9967} & \textbf{0.8476} & \textbf{0.0007} & \textbf{0.0051}\\
        ForestDML & 0.9941 & 0.8686 & 0.0021  & 0.0099  & 0.9608 & 0.8129 & 0.0026 & 0.0056 \\
        ForestDR        & $\leq 0$     & $\leq 0$     & 6.4268  & 29.0210 & 0.9581 & 0.8179 & 0.0027 & 0.0055 \\
        ForestS & 0.9771 & 0.8777 & 0.0041 & 0.0096 & 0.8243 & 0.8286 & 0.0054 & 0.0054 \\
        ForestT & 0.9793 & 0.8588 & 0.0039 & 0.0103 & 0.9454 & 0.8126 & 0.0030 & 0.0056 \\
        LinReg          & 0.9146 & 0.8646 & 0.0080  & 0.0101  & 0.6538 & 0.5599 & 0.0076 & 0.0086 \\
        LinearDML       & 0.979  & 0.8655 & 0.0040  & 0.0100  & 0.9608 & 0.8216 & 0.0026 & 0.0055 \\
        LinearDR        & $\leq 0$     & $\leq 0$     & 11.0736 & 29.5414 & 0.9589 & 0.8176 & 0.0026 & 0.0055 \\
        LinearS & 0.9146 & 0.8632 & 0.0080 & 0.0101 & 0.6538 & 0.4869 & 0.0076 & 0.0093 \\
        LinearT & 0.9181 & 0.8688 & 0.0078 & 0.0099 & 0.6395 & 0.5385 & 0.0078 & 0.0088 \\
        RF              & 0.976  & 0.8766 & 0.0042  & 0.0096  & 0.8122 & 0.8295 & 0.0056 & 0.0054\\
        \bottomrule
    \end{tabular}
\end{table*}

\begin{table*}
    \footnotesize
    \caption{Mean $R^2$ of methods estimating \glspl{cate}/\glspl{ite}, comparing estimation with datasets of size 1,000 versus 10,000.}
    \label{tab:dataset_size_results}
    \centering
    \begin{tabular}{lllllllllll}
        \\\toprule
        \multirow{3}{*}{Method} & \multicolumn{4}{c}{All Covariates} & \multicolumn{4}{c}{Hidden $\bU$} \\ 
        & \multicolumn{2}{c}{$\mP=$ GPT-2} & \multicolumn{2}{c}{$\mP=$ Llama-3-8b} & \multicolumn{2}{c}{$\mP=$ GPT-2} & \multicolumn{2}{c}{$\mP=$ Llama-3-8b} \\
        & 1,000 & 10,000 & 1,000 & 10,000 & 1,000 & 10,000 & 1,000 & 10,000 \\\midrule
        T-Only OLS & $\leq$0 & $\leq$0 & $\leq$0 & $\leq$0 & $\leq$0 & $\leq$0 & $\leq$0 & $\leq$0 \\\midrule
        BART & \textbf{0.7162} & \textbf{0.9691} & \textbf{0.5221} & \textbf{0.9214} & $\leq$0 & $\leq$0 & $\leq$0 & $\leq$0 \\
        BART-ITE & 0.7102 & 0.9344 & 0.5183 & 0.8823 & 0.0054 & 0.0185 & 0.0011 & 0.0169 \\
        CausalForest & $\leq$0 & $\leq$0 & $\leq$0 & $\leq$0 & $\leq$0 & 0.0313 & $\leq$0 & 0.1399 \\
        ForestDML & $\leq$0 & 0.6605 & $\leq$0 & 0.5551 & $\leq$0 & \textbf{0.0608} & $\leq$0 & 0.1850 \\
        ForestDR & $\leq$0 & 0.7220 & $\leq$0 & 0.5968 & $\leq$0 & 0.0503 & $\leq$0 & \textbf{0.1931} \\
        ForestS & $\leq$0 & 0.0661 & $\leq$0 & 0.1263 & 0.0026 & 0.0371 & \textbf{0.0063} & 0.0936 \\
        ForestT & 0.1031 & 0.3202 & $\leq$0 & 0.1319 & \textbf{0.0076} & 0.0366 & 0.0054 & 0.0719 \\
        LinReg & $\leq$0 & $\leq$0 & $\leq$0 & $\leq$0 & $\leq$0 & $\leq$0 & $\leq$0 & $\leq$0 \\
        LinearDML & 0.2381 & 0.5309 & 0.1491 & 0.5404 & $\leq$0 & $\leq$0 & $\leq$0 & 0.1483 \\
        LinearDR & 0.2329 & 0.5900 & 0.1562 & 0.5293 & $\leq$0 & $\leq$0 & $\leq$0 & 0.1734 \\
        LinearS & $\leq$0 & $\leq$0 & $\leq$0 & $\leq$0 & $\leq$0 & $\leq$0 & $\leq$0 & $\leq$0 \\
        LinearT & 0.0116 & 0.1470 & $\leq$0 & 0.1355 & $\leq$0 & $\leq$0 & $\leq$0 & $\leq$0 \\
        RF & $\leq$0 & 0.0327 & $\leq$0 & 0.1038 & 0.0024 & 0.0373 & 0.0060 & 0.0940 \\
        TARNet & $\leq$0 & $\leq$0 & $\leq$0 & $\leq$0 & $\leq$0 & $\leq$0 & $\leq$0 & $\leq$0 \\
        TNet & $\leq$0 & $\leq$0 & $\leq$0 & $\leq$0 & $\leq$0 & $\leq$0 & $\leq$0 & $\leq$0\\
        \bottomrule
    \end{tabular}
\end{table*}

\begin{figure*}
    \centering
    \includegraphics[width=\textwidth]{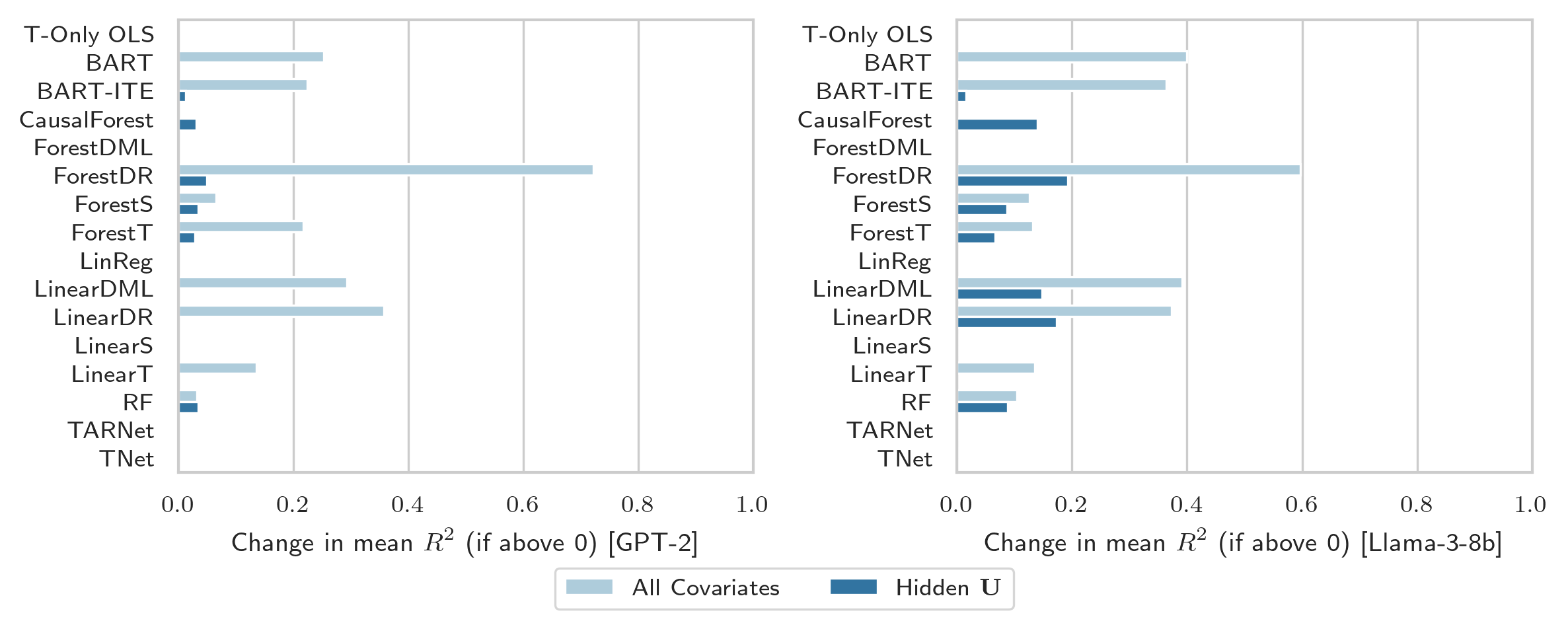}
    \caption{Change in mean $R^2$ (if above 0) of methods estimating \glspl{cate} and \glspl{ite} after a tenfold increase in dataset size from $1,000$ to $10,000$.}
    \label{fig:dataset_size_barplot}
\end{figure*}

\begin{figure*}
    \centering
    \includegraphics[width=\textwidth]{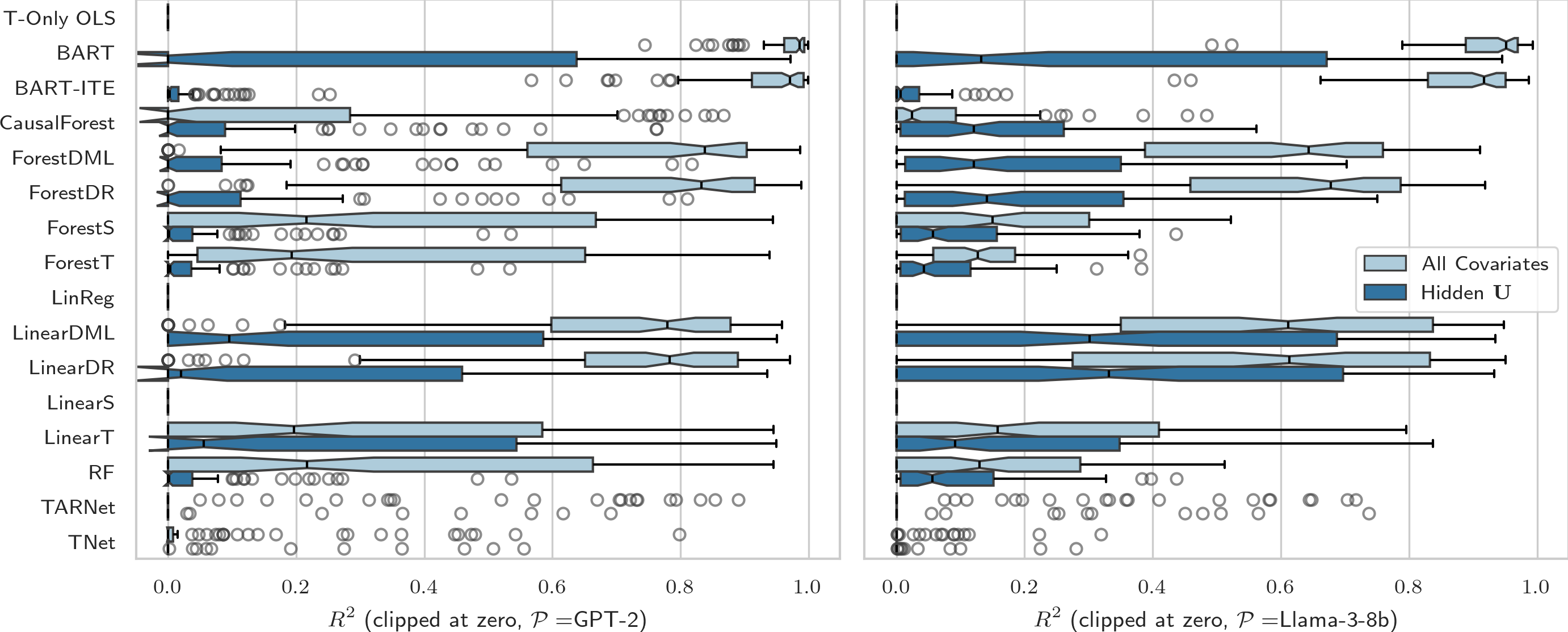}
    \caption{$R^2$ values across all methods that provide point estimates of \glspl{cate}/\glspl{ite} for datasets of size 10,000 generated by GPT-2 (left) and Llama-3-8b (right).}
    \label{fig:cate_r2_boxplots_10k}
\end{figure*}

\begin{figure*}
    \centering
    \includegraphics[width=\textwidth]{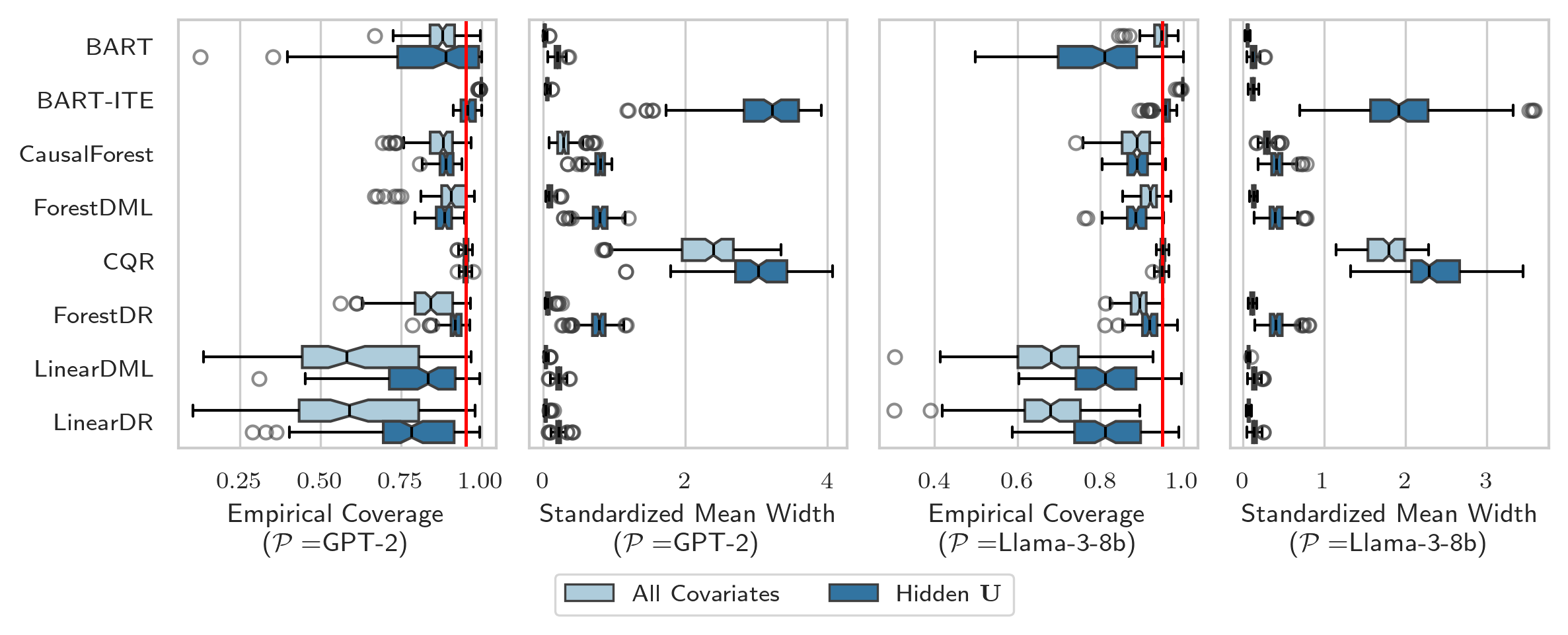}
    \caption{\gls{cate}/\gls{ite} empirical coverage ($\alpha = 0.05$) and interval width (in outcome standard deviation units) for methods that provide intervals. Nominal coverage of 95\% is indicated by the red line. Shown for datasets of size 10,000 generated by GPT-2 (left) and Llama-3-8b (right).}
    \label{fig:cate_coverage_boxplots_10k}
\end{figure*} 

\begin{figure*}
    \centering
    \includegraphics[width=\textwidth]{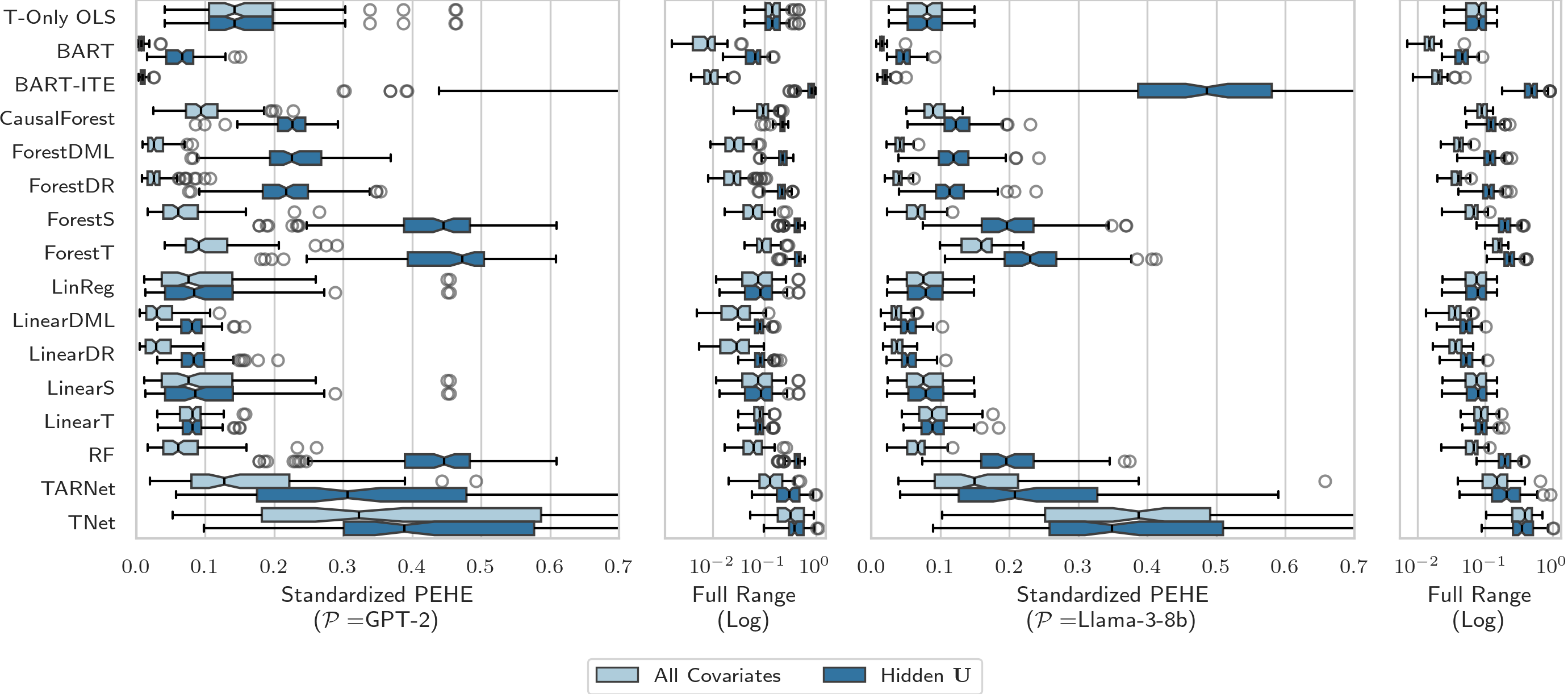}
    \caption{PEHE across all methods that provide point estimates of \glspl{cate}/\glspl{ite}, shown in standard deviation units of the outcome $\ty$. Results shown for datasets of size 10,000 generated by GPT-2 (left) and Llama-3-8b (right).}
    \label{fig:pehe_boxplots_10k}
\end{figure*}

\begin{figure*}
    \centering
    \includegraphics[width=\textwidth]{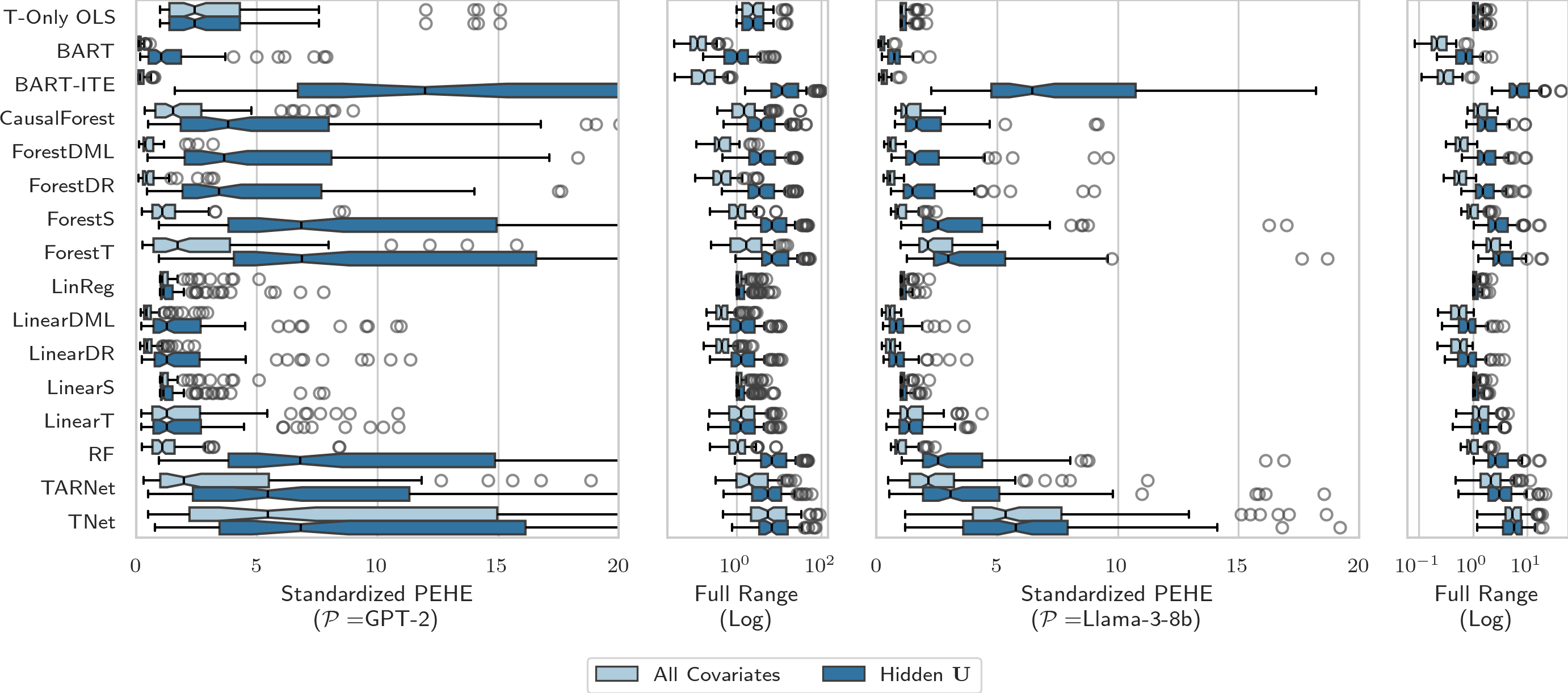}
    \caption{PEHE across all methods that provide point estimates of \glspl{cate}/\glspl{ite}, shown in units of \gls{ite} standard deviation. Results shown for datasets of size 10,000 generated by GPT-2 (left) and Llama-3-8b (right).}
    \label{fig:pehe_boxplots_sd_ite_10k}
\end{figure*}

\begin{figure*}
    \centering
    \includegraphics[width=\textwidth]{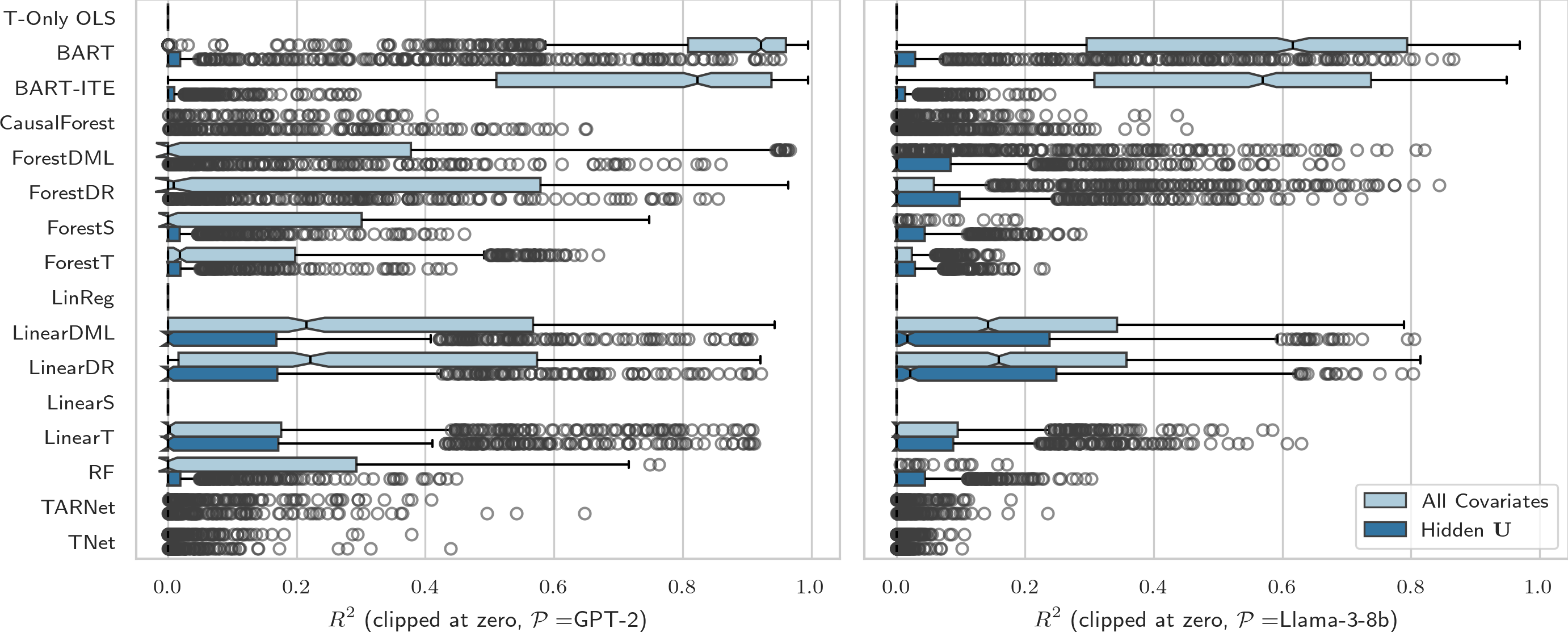}
    \caption{$R^2$ values across all methods that provide point estimates of \glspl{cate}/\glspl{ite} for datasets of size 1,000 generated by GPT-2 (left) and Llama-3-8b (right).}
    \label{fig:cate_r2_boxplots}
\end{figure*}

\begin{figure*}
    \centering
    \includegraphics[width=\textwidth]{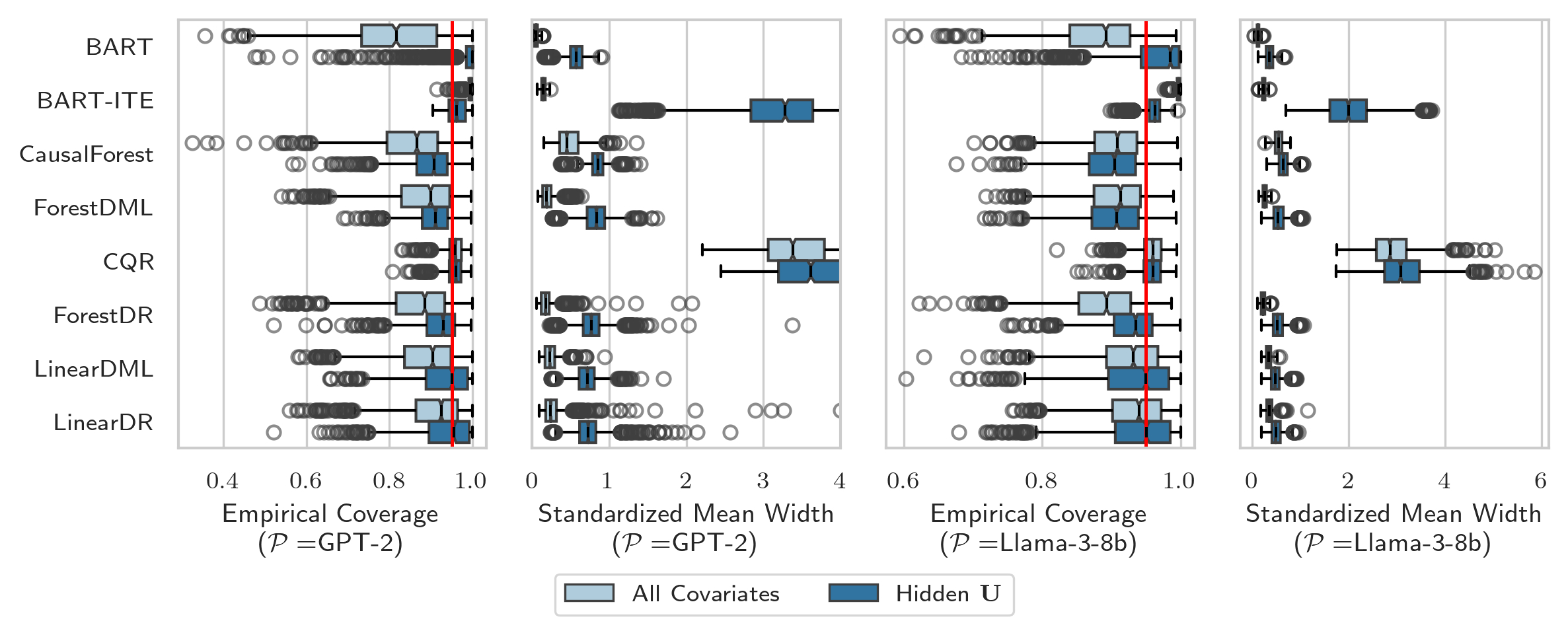}
    \caption{\gls{cate}/\gls{ite} empirical coverage ($\alpha = 0.05$) and interval width (in outcome standard deviation units) for methods that provide intervals. Nominal coverage of 95\% is indicated by the red line. Shown for datasets of size 1,000 generated by GPT-2 (left) and Llama-3-8b (right).}
    \label{fig:cate_coverage_boxplots}
\end{figure*} 

\begin{figure*}
    \centering
    \begin{minipage}{0.49\textwidth}
        \centering
        \includegraphics[width=\textwidth]{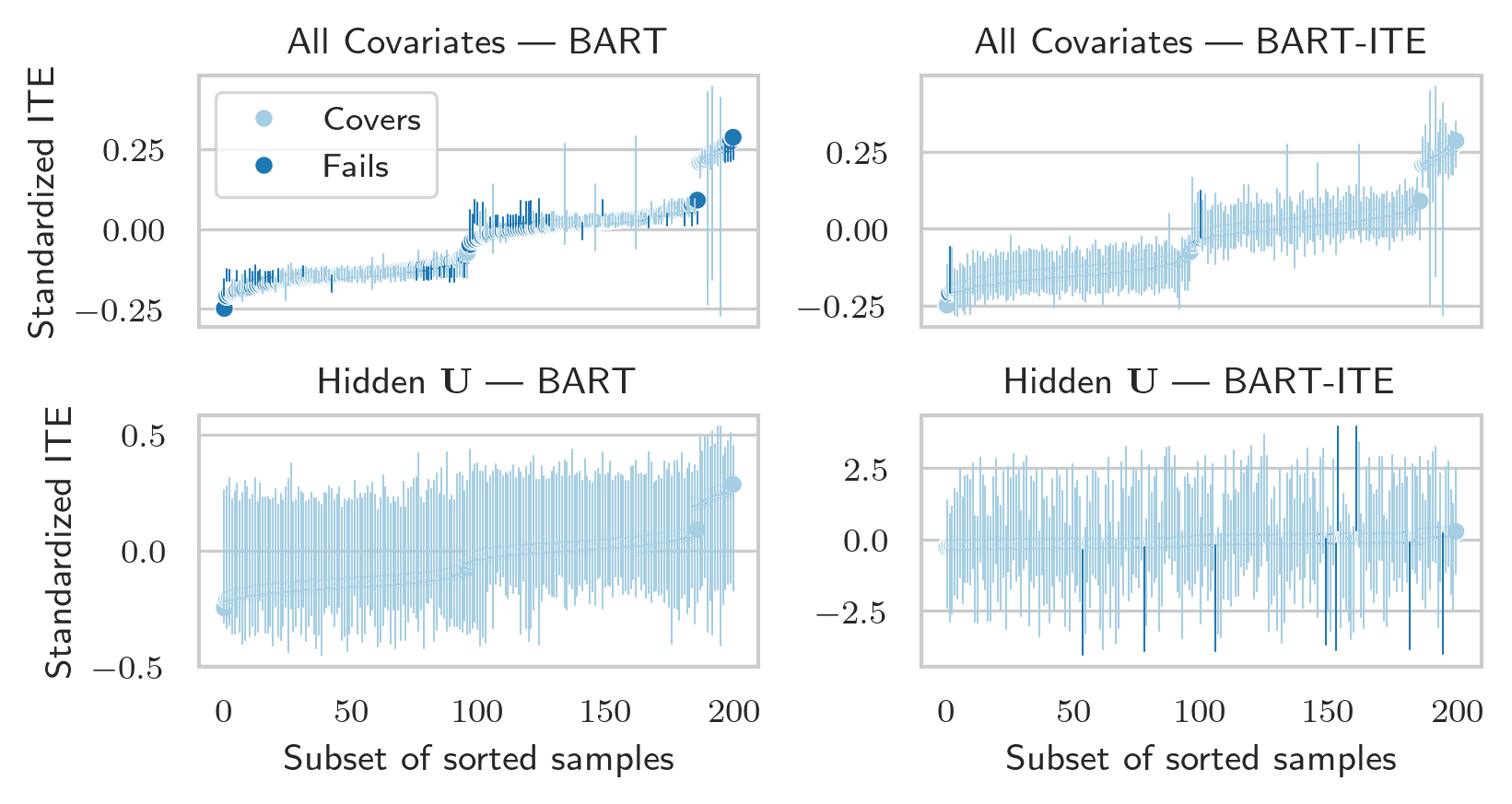}\\
        \customlabel{fig:ite_intervals_1}{a}{\ref{fig:ite_intervals_1}}
    \end{minipage}
    \begin{minipage}{0.49\textwidth}
        \centering
        \includegraphics[width=\textwidth]{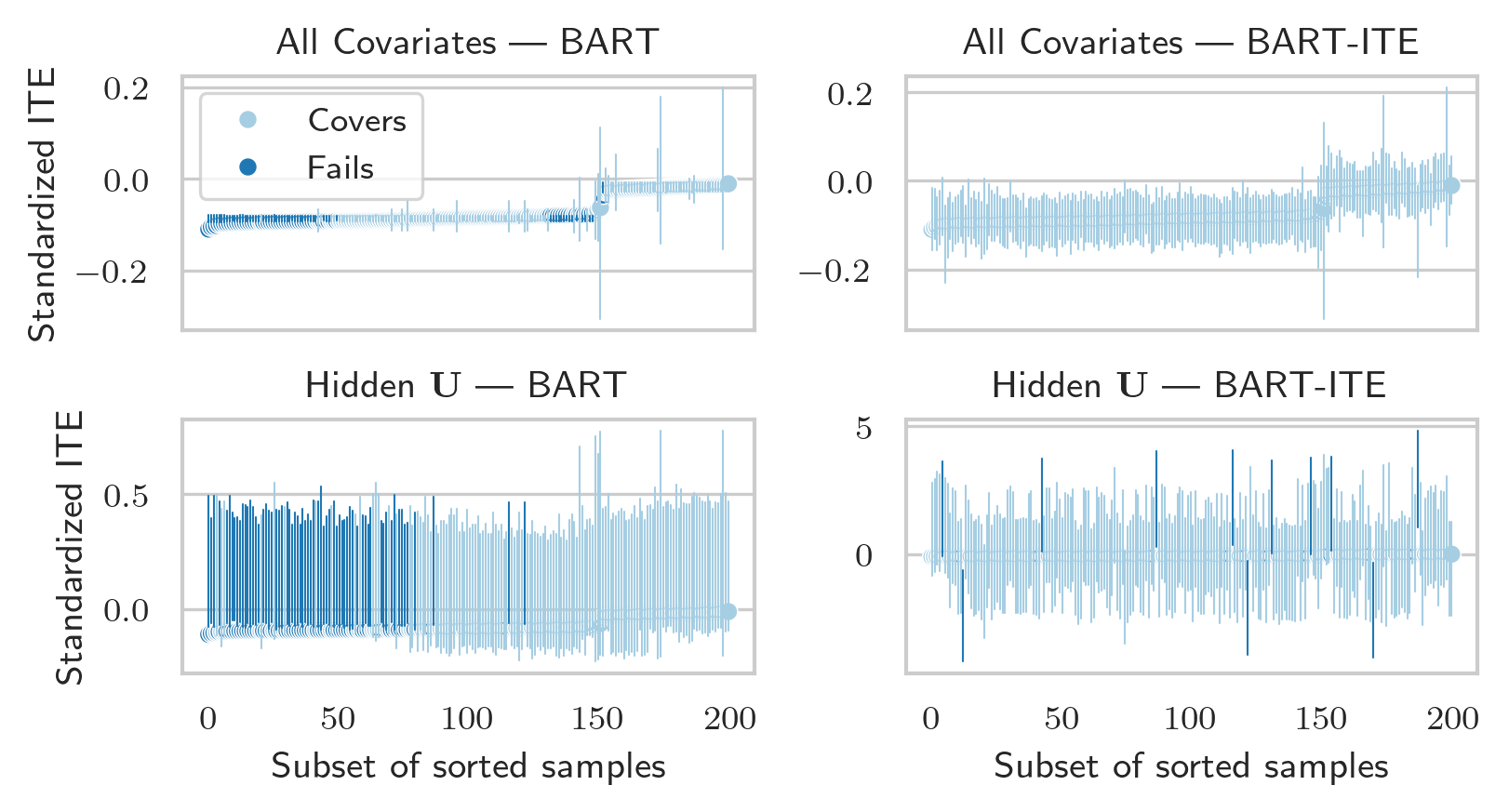}\\
        \customlabel{fig:ite_intervals_2}{b}{\ref{fig:ite_intervals_2}}
    \end{minipage}
    \caption{Interval estimates from BART versus BART-ITE across two example datasets of size 1,000, (a) and (b).}
    \label{fig:ite_intervals}
\end{figure*}

\Cref{fig:bcancer_dataset_plots} shows plots of the features (\ref{fig:bcancer_hist}), propensity scores (\ref{fig:bcancer_propensity}), categorical \glspl{ite} (\ref{fig:bcancer_ites_binary}), and continuous \glspl{ite} (\ref{fig:bcancer_ites}) for a single generated dataset using Llama-3-8b. Because the outcome $\ty$ has $|\Omega_{\ty}| = 4$ possible values, we can consider several possible outcomes, including the observed outcome (categorical), log probabilities for each outcome value (continuous), or probabilities for each outcome value (continuous). This creates, in effect, nine possible targets for each dataset. For benchmarking purposes, we find using probabilities and/or log probabilities as the outcome to be the most useful --- there is frequently an effect (even at the individual level) in probability space, even if the sampled outcomes do not change. Comparing Figure~\ref{fig:bcancer_ites_binary} to Figure~\ref{fig:bcancer_ites} demonstrates this, where for the 400 or so observations where the categorical \gls{ite} is zero, the continuous \gls{ite} is instead nonzero.

Figure~\ref{fig:bcancer_ites} also demonstrates that we are able to satisfy our main criterion for meaningful benchmark: the observational distribution $P_{\ty \mid \tt=t}^{\fB}$ and interventional distribution $P_{\ty}^{\fB; \doTt}$ are different enough that the \gls{sate} and the observed mean difference in outcomes between the treatment and control group are not only different in value, but \emph{they also disagree in sign}. This is particularly meaningful in a causal inference setting --- the treatment appears to lower the outcome, when in fact, its effect is to increase the outcome.

\section{Additional estimation results}\label{appendix:additional_results}

\subsection{\glsentryshort{ate} results}\label{appendix:ate_results}

For all implementations that directly support \gls{ate} estimation, we report the $R^2$ and root-mean-squared-error (RMSE) across the 1000 datasets for each language model in two settings, using $\text{logP}(\ty = 0)$ as the outcome. The first setting is with estimation using all 14 covariates (all 12 confounders, the treatment, and the outcome). This is denoted \textbf{All Cov.} in \Cref{tab:bcancer_ate_results}. The second setting is with the variables $\bU=\{\tu_1, \tu_2, \tu_3, \tu_4\} =$ \{\texttt{age}, \texttt{medical conditions}, \texttt{medication}, \texttt{menopausal status}\} hidden, denoted \textbf{Hidden}. \Cref{tab:bcancer_ate_results} shows that \gls{ate} estimation is more or less challenging depending on which language model is used. In this case, Llama-3-8b produces \glspl{ate} that are more challenging to estimate, with the exception of GPT-2 for the doubly robust methods, whose $R^2$ and RMSE suffer significantly due to several large outlying estimates. Across all methods, performance tends to drop significantly in the `Hidden' setting, suggesting that $\bU$ are indeed hidden confounders. Across methods, BART shows the strongest performance in all settings in \Cref{tab:bcancer_ate_results}.

\subsection{Comparison of dataset size 1,000 and 10,000}

\Cref{tab:dataset_size_results,fig:dataset_size_barplot} show that \gls{cate} and \gls{ate} estimation remain difficult even after a tenfold increase in dataset size (from $N=1,000$ to $N=10,000$), especially in the Hidden $\bU$ setting. Across estimation methods, performance tends to increase as sample size increases, especially if the method originally achieved $R^2$ above zero with $N=1,000$. In other words, methods that do reasonably well at $N=1,000$ show improvement with more data, as we would expect. However, several methods struggle in both settings, even with ten times more data. For example, TARNet, TNet, and CausalForest still remain unstable and inaccurate in both the All Covariates and the Hidden $\bU$ settings across both sample sizes. Overall, these results indicate that CATE and ITE estimation in this case are not challenging due only to small sample sizes. This is useful to know, especially when we consider that corresponding real-world use-cases often deal with even smaller sample sizes.

Additional results for each dataset size are shown individually in the following figures. \Cref{fig:cate_r2_boxplots_10k,fig:cate_coverage_boxplots_10k} show $R^2$ and coverage results on datasets of size 10,000. These correspond to the same figures in the main text, but now showing both GPT-2 and Llama-3-8b, allowing for comparison across models. \Cref{fig:pehe_boxplots_10k} shows the same setting using standardized Precision in Estimating Heterogeneous Effects (PEHE) \cite{Hill2011BayesianNM}, which is the RMSE of the \gls{cate} predictions across the different observed values of $x$, i.e., 
$$\text{PEHE}_j=\sqrt{\frac{1}{n} \sum_{i=1}^{n}(\hat{\tau}_i - \tau_i)^2}$$ 
for a dataset $D_j = \{\bu_i, \bx_i, t_i, y_i\}_{i=1}^{n}$ where $\hat{\tau}_i$ is the estimated \gls{ite} for unit $i$ and $\tau_i$ is the true \gls{ite}. 
We standardize PEHE using the empirical standard deviation $\hat{\sigma}_{j}$ of the outcomes $\{y_i\}_{i=1}^{n}$ in each dataset, i.e., 
$$\text{(Standardized PEHE)}_j =\sqrt{\frac{1}{n \cdot \hat{\sigma}^2_{j}} \sum_{i=1}^{n}(\hat{\tau}_i - \tau_i)^2}.$$ \Cref{fig:pehe_boxplots_sd_ite_10k} shows the same metric standardized instead using the (much smaller) standard deviation of the \gls{ite}.

Results in the case of dataset size $1,000$ show similar trends to those in the size $10,000$ setting. \Cref{fig:cate_r2_boxplots} shows $R^2$ values clipped at zero across all methods that provide point estimates for \glspl{cate}. When all covariates are observed, BART does significantly better explaining \gls{cate} variation, followed by DML and DR with much lower averages, much like the size 10,000 case. Similarly, \gls{cate} estimation becomes much more challenging for all methods with hidden $\bU$. The difference in effect estimation difficulty between Llama-3-8b and GPT-2 is also more noticeable for \glspl{cate} than it was for \glspl{ate}. 
Overall, some methods show more instability in the dataset size 1,000 case than in the size 10,000 case, as expected with less data.

\Cref{fig:cate_coverage_boxplots} shows empirical coverage results in the dataset size 1,000 case for all estimators that provide intervals. Similar to the size 10,000 case, empirical coverage is under nominal for all methods that target \gls{cate} in the setting with all covariates. Hidden $\bU$ generally increases uncertainty, bringing the DR methods and LinearDML median coverage near nominal. Interestingly, BART for \gls{cate} achieves higher median coverage of the \gls{ite} than BART-ITE, but with a much larger tail of poor coverage. BART-ITE, by contrast, has much less variable coverage in the \gls{ite} setting, but at the cost of much wider intervals. \Cref{fig:ite_intervals} shows intervals for BART targeting the CATE versus BART-ITE across two example datasets of size 1,000, demonstrating that, as in the size 10,000 case, the tighter intervals of BART targeting the CATE can be overconfident with variable coverage, while the wider intervals of BART-ITE are too wide to be useful. 

\end{document}